\documentclass{article}

\usepackage{microtype}
\usepackage[pdftex]{graphicx}
\usepackage{subcaption}
\usepackage{booktabs} 
\usepackage{bm, amsmath, amssymb, amsthm}
\usepackage{tabularx}
\usepackage{makecell}
\usepackage{threeparttable}
\usepackage[colorinlistoftodos]{todonotes}
\usepackage{hyperref}

\usepackage[arxiv]{icml2020}

\def\x{{\bm{x}}}
\def\y{{\bm{y}}}
\def\z{{\bm{z}}}
\def\data{{\mathcal{D}}}

\icmltitlerunning{Inverse Graphics GAN: Learning to Generate 3D Shapes from Unstructured 2D Data}

\begin{document}

\twocolumn[
\icmltitle{Inverse Graphics GAN: \\  Learning to Generate 3D Shapes from Unstructured 2D Data}




\begin{icmlauthorlist}
\icmlauthor{Sebastian Lunz}{INTERN}
\icmlauthor{Yingzhen Li}{MSR}
\icmlauthor{Andrew Fitzgibbon}{MSR}
\icmlauthor{Nate Kushman}{MSR}
\end{icmlauthorlist}

\icmlaffiliation{MSR}{Microsoft Research}
\icmlaffiliation{INTERN}{Work done during an internship at Microsoft Research}

\icmlcorrespondingauthor{Sebastian Lunz}{sl767@cam.ac.uk}
\icmlcorrespondingauthor{Nate Kushman}{nate@kushman.org}

\icmlkeywords{Inverse Graphics, GANs}

\vskip 0.3in
]



\printAffiliationsAndNotice{}  

\begin{abstract}

Recent work has shown the ability to learn generative models for 3D shapes from only unstructured 2D images.  However, training such models requires differentiating through the rasterization step of the rendering process, therefore past work has focused on developing bespoke rendering models which smooth over this non-differentiable process in various ways.
Such models are thus unable to take advantage of the photo-realistic, fully featured, industrial renderers built by the gaming and graphics industry.  In this paper we introduce the first scalable training technique for 3D generative models from 2D data which utilizes an off-the-shelf non-differentiable renderer. To account for the non-differentiability, we introduce a proxy neural renderer to match the output of the non-differentiable renderer. We further propose discriminator output matching to ensure that the neural renderer learns to smooth over the rasterization appropriately. We evaluate our model on images rendered from our generated 3D shapes, and show that our model can consistently learn to generate better shapes than existing models when trained with exclusively unstructured 2D images.
\end{abstract}

\section{Introduction}

Generative adversarial networks (GANS) have produced impressive results on 2D image data~\cite{styleGAN, bigGAN}.  Many visual applications, such as gaming, require 3D models as inputs instead of just images, however, and directly extending existing GAN models to 3d, requires access to 3D training data~\cite{wu2016learning, riegler2017octnet}.  This data is expensive to generate and so exists in abundance only for only very common classes.
Ideally we'd like to be able to learn to generate 3D models while training with only 2D image data which is much more widely available and much cheaper and easier to obtain.

\begin{figure}[]
    \centering

    \includegraphics[trim={3cm 4cm 3cm 4cm}, clip, width=.15\textwidth]{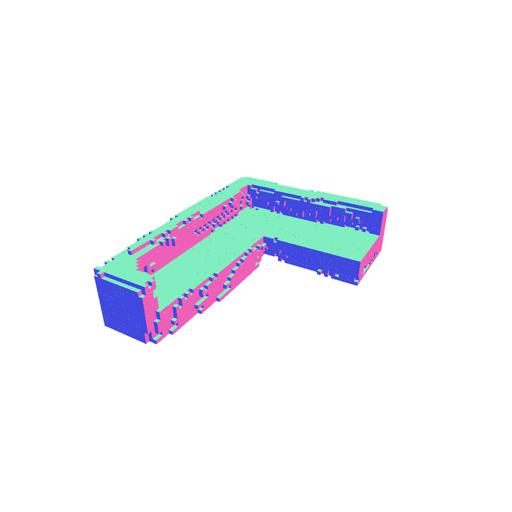}
    \includegraphics[trim={3cm 3cm 3cm 3cm}, clip, width=.15\textwidth]{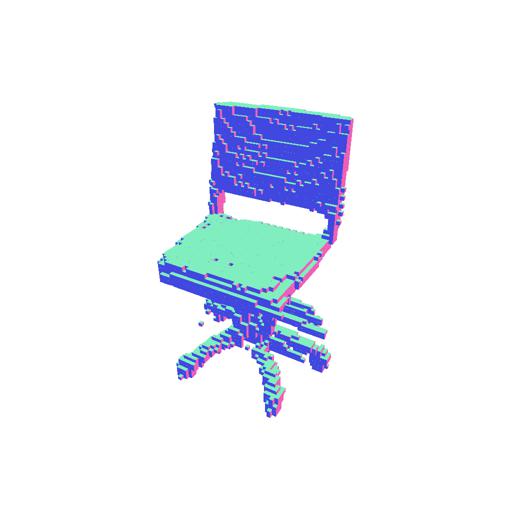}
    \includegraphics[trim={3cm 4cm 3cm 4cm}, clip, width=.15\textwidth]{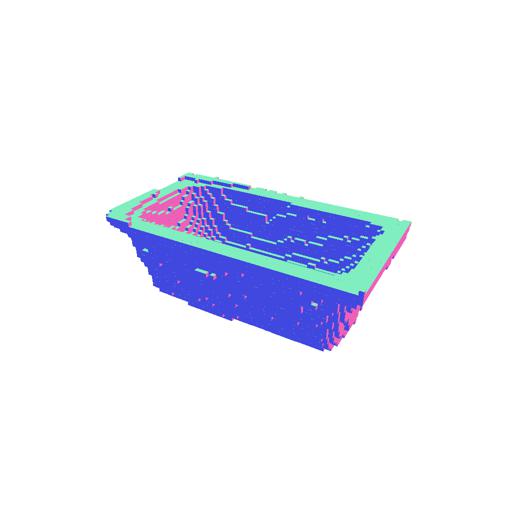}

\caption{\label{fig:Exposition} 3D shapes generated by training IG-GAN on unstructured 2D images rendered from three ShapeNet classes.}
    
\end{figure}

Training with only 2D data allows us to use any 3D representation.  Our interest is in creating 3D models for gaming applications which typically rely on 3D meshes, but direct mesh generation is not ammenable to generating arbitary topologies since most approaches are based on deforming a template mesh.  So we instead choose to work with voxel representations because they can represent arbitrary topologies, can easily be converted to meshes using the marching cubes algorithm, and can be made differentiable by representing the occupancy of each voxel by a real number $\in [0,1]$ which identifies the probability of voxel occupancy.

In order to learn with an end-to-end differentiable model, we need to differentiate through the process of rendering the 3D model to a 2D image, but the rasterization step in rendering is inherently non-differentiable.  As a result, past work on 3D generation from 2D images has focused on differentiable renderers which are hand built from scratch to smooth over this non-differentiable step in various ways.  However, standard photo realistic industrial renderers created by the gaming industry (e.g. UnReal Engine, Unity) are not differentiable, and so with such methods we cannot use these renderers, and must rely instead on the simple differentiable renderers build by the research community.
In particular two aspects of the rendering process are non-differentiable: (1) the rasterization step inside of the renderer is inherently non-differentiable as a result of occlusion and (2) sampling the continuous voxel grid to generate a mesh is also not differentiable.  This is second step is required because typical industrial renderers take a mesh as input and we can easily convert a binary voxel grid to a mesh, but continuous voxel inputs do not have a meaningful mesh representation.  So rendering a continuous voxel grid using an off-the-shelf renderer requires first sampling a binary voxel grid from the distribution defined by the continuous voxel grid.



In this paper we introduce the first {\em scalable} training technique for 3D generative models from 2D data which utilises an off-the-shelf non-differentiable renderer.  Examples of the result of our method can be seen in Figure~\ref{fig:Exposition}.  Key to our method is the introduction of a proxy neural renderer based on the recent successes of neural rendering \cite{rendernet} which directly renders the continuous voxel grid generated by the 3D generative model. It addresses the two challenges of the non-differentiability of the off-the-shelf render as follows:

\noindent\textbf{Differentiate through the Neural Renderer:}  The proxy neural renderer is trained to match the rendering output of the off-the-shelf renderer given a 3D mesh input.  This allows back-propagation of the gradient from the GAN discriminator through the neural renderer to the 3D generative model, enabling training using gradient descent.

\noindent\textbf{Discriminator Output Matching:}  In order to differentiate through the voxel sampling step we also train the proxy neural renderer using a novel loss function which we call {\em discriminator output matching}.  
This accounts for the fact that the neural renderer can only be trained to match the off-the-shelf renderer for binary inputs, which leaves it free to generate  arbitrary outputs for the (typically) non-binary voxel grids created by the generator. We constrain this by computing the discriminator loss of an image rendered by the neural renderer when passed through the discriminator. This loss is matched to the average loss achieved by randomly thresholding the volume, rendering the now binary voxels with the off-the-shelf renderer, and passing the resulting image through the discriminator.
This addresses the instance-level non-differentiability issue and instead targets the differentiable loss defined on the population of generated discrete 3D shapes, forcing the neural renderer to generate images which represent the continuous voxel grids as smoothly interpolation between the binary choice from the perspective of the discriminator.

We evaluate our model on a variety of synthetic image data sets generated from 3D models in ShapeNet \cite{ShapeNet} as well as the natural image dataset of Chanterelle mushrooms introduced in \citet{platosCave}, and show that our model generates 3D meshes whose renders generate improved 2D FID scores compared to both~\citet{platosCave} and a 2D GAN baseline.

\section{Related Work}

\noindent\textbf{Geometry Based Approaches (or 3D Reconstruction):} Reconstructing the underlying 3D scene from only 2D images has been one of the long-standing goals of computer vision.  Classical work in this area has focused on geometry based approaches in the single instanced setting where the goal was only to reconstruct a single 3D object or scene depicted in one or more 2D images \cite{bleyer2011patchmatch, de1999poxels, broadhurst2001probabilistic, galliani2015massively, kutulakos2000theory, prock1998towards, schonberger2016pixelwise, seitz2006comparison, seitz1999photorealistic}.  This early work was not learning based, however, and so was unable to reconstruct any surfaces which do not appear in the image(s).

\noindent\textbf{Learning to Generate from 3D Supervision:}
Learning-based  3D  reconstruction techniques use a training set of samples to learn a distribution over object shapes. 
Much past work has focused on the simplest learning setting in which we have access to full 3D supervision.  This includes work on generating voxels~\cite{brock2016generative, choy20163d, riegler2017octnet, wu2016learning,xie2019pix2vox}, generating point-clouds~\cite{achlioptas2017learning, fan2017point, jiang2018gal, yang2019pointflow, achlioptas2017learning, li2018point}, generating meshes~\cite{groueix2018papier, pan2019deep, wang2018pixel2mesh} and generating implicit representations~\cite{atzmon2019controlling, chen2019learning, genova2019learning,huang2018deep, mescheder2019occupancy, michalkiewicz2019implicit, park2019deepsdf, saito2019pifu, xu2019disn}.  Creating 3D training data is much more expensive, however, because it requires either skilled artists or a specialized capture setup.  So in contrast to all of this work we focus on learning only from unstructured 2D image data which is more readily available and cheaper to obtain.

\noindent\textbf{Learning to Generate from 2D Supervision:}
Past work on learning to generate 3D shapes by training on only 2D images has mostly focused on differentiable renderers.  We can categorize this work based on the representation used. Mesh techniques~\cite{kanazawa2018learning, chen2019learning, genova2018unsupervised, henderson2019learning} are based on deforming a single template mesh or a small number of pieces~\cite{henderson2019learning}, while \citet{loper2014opendr} and \citet{palazzi2018end} use only a low-dimensional pose representation, so neither is  amenable to generating arbitrary topologies.  Concurrent work on implicit models~\cite{niemeyer2019differentiable, liu2019learning} can directly learn an implicit model from 2D images without ever expanding to another representation, but these methods rely on having camera intrinsics for each image, which is usually unavailable with 2D image data. Our work instead focuses on working with unannotated 2D image data. 

The closet work to ours uses voxel representations~\cite{PrGAN, platosCave}.  Voxels can represent arbitrary topologies and can easily be converted to a mesh using the marching cubes algorithm.  Furthermore,  although it is not a focus of this paper, past work has shown that the voxel representation can be scaled to relatively high resolutions through the use of sparse representations~\cite{riegler2017octnet}.
\citet{PrGAN} employs a visual hull based differential renderer that only considers a smoothed version of the object silhouette, while \citet{platosCave} relies on a very simple emission-absorption based lighting model.  As we show in the results, both of these models struggle to take advantage of lighting and shading information which reveals surface differences, and so they struggle to correctly represent concavities like bathtubs and sofas.

In contrast to all previous work, our goal is to be able to take advantage of fully-featured industrial renderers which included many advanced shading, lighting and texturing features.  However these renderers are typically not built to be differentiable, which is challenging to work with in machine learning pipelines. The only work we are aware of which uses an off-the-shelf render for 3D generation with 2D supervision is~\citet{rezende2016unsupervised}.  In order to differentiate through the rendering step they use the REINFORCE gradients ~\cite{williams1992simple}.  However, REINFORCE scales very poorly with number of input dimensions,allowing them to show results on simple meshes only. In contrast, our method scales {\em much} better since dense gradient information can flow through the proxy neural renderer.

\noindent\textbf{Neural Rendering}
With the success of 2D generative models, it has recently become popular to skip the generation of an explicit 3D representation  {\em Neural Rendering} techniques focus only on simulating 3D by using a neural network to generate 2D images directly from a latent space with control over the camera angle~\cite{eslami2018neural, HoloGAN, sitzmann2019scene} and properties of objects in the scene~\cite{liao2019towards}.
In contrast, our goal is to generate the 3D shape itself, not merely controllable 2D renders of it.  This is important in circumstances like gaming where the underingly rendering framework is may be fixed, or where we need direct access to the underlying 3D shape itself, such as in CAD/CAM applications.  We do however build directly on RenderNet~\citet{rendernet} which is a neural network that can be trained to generate 2D images from 3D shapes by matching the output of an off-the-shelf renderer.  

\noindent\textbf{Differentiating Through Discrete Decisions}
Recent work has looked at the problem of differentiating through discrete decisions.~\citet{maddison2016concrete} and~\citet{jang2016categorical} consider smoothing over the discrete decision and~\citet{tucker2017rebar} extends this with sampling to debias the gradient estimates.  In section~\ref{sec:method} we discuss why these methods cannot be applied in our setting.  \citet{liao2018deep} discusses why we cannot simply differentiate through the Marching Cubes algorithm, and also suggests using continuous voxel values to generate a probability distribution over 3D shapes.  However in their setting they have ground truth 3D data so they directly use these probabilities to compute a loss and do not have to differentiate through the voxel sampling process as we do when training from only 2D data.

\begin{figure*}[t]
    \centering
    \includegraphics[width=1\linewidth]{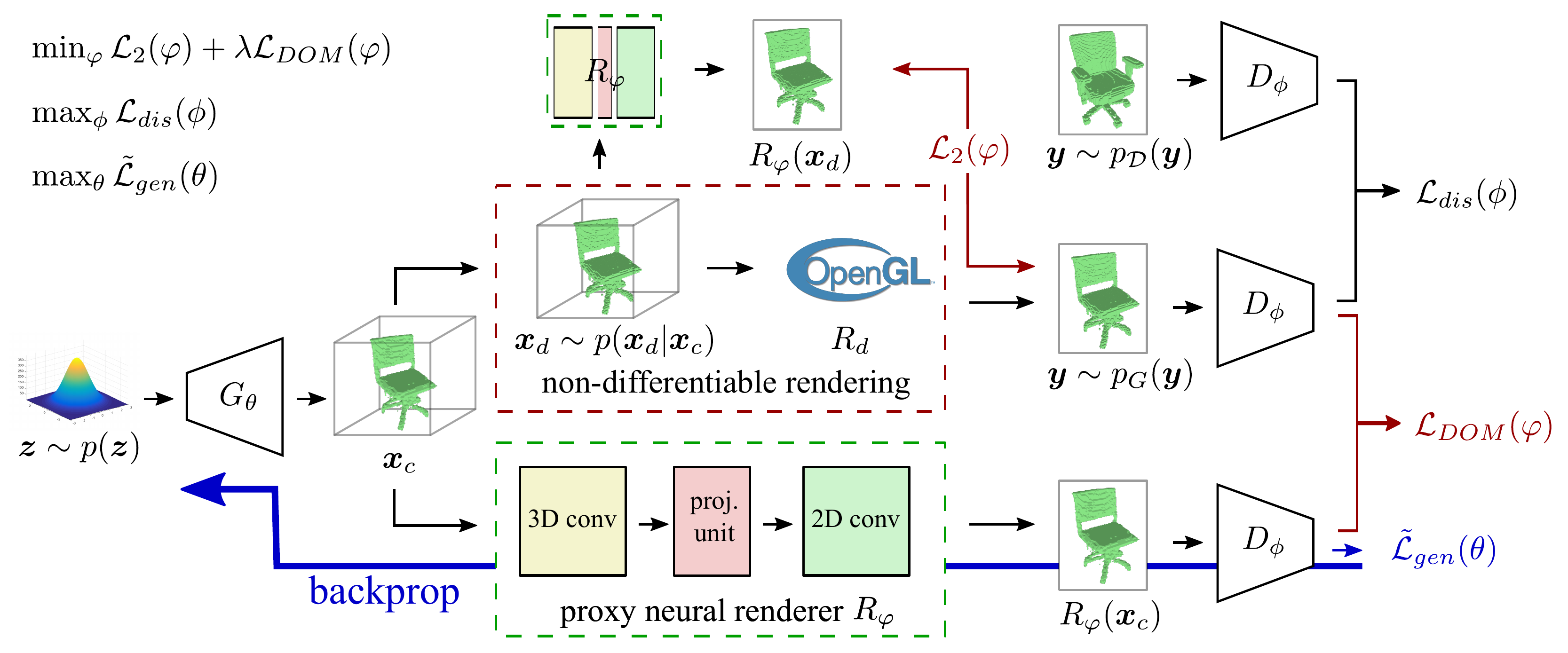}
    \caption{The architecture and training setup for IG-GAN. }
    \label{fig:method_visualisation}
\end{figure*}

\section{IG-GAN}
\label{sec:method}
We wish to train a generative model for 3D shapes such that rendering these shapes with an off-the-shelf renderer generates images that match the distribution of 2D training image dataset.
The generative model $G_{\theta}(\cdot)$ takes in a random input vector $\z \sim p(\z)$ and generate a continuous voxel representation of the 3D object $\x_c = G_{\theta}(\z)$. Then the voxels $\x_c$ are fed to a \emph{non-differentiable} renderering process,
where the voxels first are thresholded to discrete values $\x_d \sim p(\x_d | \x_c)$, then the discrete-value voxels $\x_d$ are renderred using the off-the-shelf renderer (e.g.~OpenGL) $\y = R_d(\x_d)$. In summary, this generating process samples a 2D image $\y \sim p_G(\y)$ as follows:
\begin{equation}
\hspace{-0.7em}
\resizebox{0.92\hsize}{!}{$
\begin{aligned}  
  \x_c &\sim p_G(\x_c) \Leftrightarrow \z \sim p(\z), \x_c = G_{\theta}(\z), \\
    \y &\sim p_G(\y) \Leftrightarrow \x_c \sim p_G(\x_c), \x_d \sim p(\x_d | \x_c), \y = R_d(\x_d).
\end{aligned}$
  }
\label{eq:full_generation_process}
\end{equation}
Like many GAN algorithms, a discriminator $D_{\phi}$ is then trained on both images sampled from the 2D data distribution $p_{\data}(\y)$ and generated images sampled from $p_G(\y)$. We consider maximising e.g.~the classification-based GAN cross-entropy objective when training the discriminator
\begin{equation}
\hspace{-0.9em}
\resizebox{0.95\hsize}{!}{$
    \max_{\phi} \mathcal{L}_{\text{dis}}(\phi) := \mathbb{E}_{p_{\mathcal{D}}(\y)} \left[ \log D_{\phi}(\y) \right] + \mathbb{E}_{p_{G}(\y)} \left[ \log (1 - D_{\phi}(\y)) \right].$
}
\end{equation}

A typical GAN algorithm trains the generator $G_{\theta}(\cdot)$ by maximising a loss defined by the discriminator, e.g.~
\begin{equation}
    \max_{\theta} \mathcal{L}_{\text{gen}}(\theta) := \mathbb{E}_{p_{G}(\y)} \left[ \log D_{\phi}(\y) \right].
\label{eq:generator_population_loss}
\end{equation}
For 2D image GAN training, optimisation is usually done by gradient descent, where the gradient is computed via the reparameterisation trick \citep{salimans2013fixed,kingma2014vae,rezende2014vae}. Unfortunately, the reparameterisation trick is not applicable in our setting since the generation process (\ref{eq:full_generation_process}) involves sampling discrete variable $\x_d$ thus non-differentiable. 
An initial idea to address this issue would be to use the REINFORCE gradient estimator \citep{williams1992simple} for the generative model loss (\ref{eq:generator_population_loss}):
\begin{equation}
\small
    \nabla_{\theta} \mathcal{L}_{\text{gen}}(\theta) = \mathbb{E}_{p_G(\x_c)} \left[ \mathbb{E}_{p_(\x_d | \x_c)} \left[ \log D_{\phi}(R_d(\x_d)) \right]  S_G(\x_c) \right],
\label{eq:reinforce_gradient_exact}
\end{equation}
with $S_G(\x_c) = \nabla_{\theta} \log p_G(\x_c)$ short-hands the score function of $p_G(\x_c)$. Here the expectation term $\mathbb{E}_{p_(\x_d | \x_c)} \left[ \log D_{\phi}(R_d(\x_d)) \right] $ is often called the ``reward'' of generating $\x_c \sim p_G(\x_c)$. Intuitively, the gradient ascent update using (\ref{eq:reinforce_gradient_exact}) would encourage the generator to generate $\x_c$ with high reward, thus fooling the discriminator.
But again REINFORCE is not directly applicable as the score function $S_G(\x_c)$ is intractable (the distribution $p_G(\x_c)$ is implicitly defined by neural network transformations of noise variables). A second attempt for gradient approximation would replace the discrete sampling step $\x_d \sim p(\x_d | \x_c)$ with continuous value approximations \citep{jang2016categorical,maddison2016concrete}, with the hope that this enables the usage of the reparameterisation trick. However, the off-the-shelf renderer is treated as black-box so we do not assume the back-propagation mechanism is implemented there. Even worse, the discrete variable sampling step is necessary as off-the-shelf renderer can only work with discrete voxel maps. Therefore the instance-level gradient is not defined on the discrete voxel grid, and gradient approximation methods based on continuous relaxations cannot be applied neither.

To address the non-differentiability issues, we introduce a proxy neural renderer $R_{\varphi}(\cdot)$ as a pathway for back-propagation in generator training. In detail, we define a neural renderer $\tilde{\y} = R_{\varphi}(\x_c)$ which directly renders the continuous voxel representation $\x_c$ into the 2D image $\tilde{\y}$. 
To encourage realistic renderings that are closed to the results from the off-the-shelf renderer, the neural renderer is trained to minimise the $\ell_2$ error of rendering on discrete voxels:
\begin{equation}
    \mathcal{L}_{2}(\varphi) = \mathbb{E}_{p_G(\x_c)p(\x_d | \x_c)}  \left[ || R_{\varphi}(\x_d) - R_d(\x_d) ||^2_2 \right] .
\end{equation}
If the neural renderer matches closely with the off-the-shelf renderer on rendering discrete voxel grids, then we can replace the non-differentiable renderer $R_{d}(\cdot)$ in (\ref{eq:reinforce_gradient_exact}) with the neural renderer $R_{\varphi}(\cdot)$:
\begin{equation}
\hspace{-0.7em}
\resizebox{0.92\hsize}{!}{$
\begin{aligned}
    \nabla_{\theta} \mathcal{L}_{\text{gen}}(\theta) &= \mathbb{E}_{p_G(\x_c)} \left[ \mathbb{E}_{p_(\x_d | \x_c)} \left[ \log D_{\phi}(R_{d}(\x_d)) \right] S_G(\x_c) \right] \\
    &\approx \mathbb{E}_{p_G(\x_c)} \left[ \mathbb{E}_{p_(\x_d | \x_c)} \left[ \log D_{\phi}(R_{\varphi}(\x_d)) \right] S_G(\x_c) \right].
\end{aligned}$
}
\end{equation}
The intractability of $S_G(\x_c)$ remains to be addressed. Notice that the neural renderer can take in both discrete and continuous voxel grids as inputs, therefore the instance-level gradient $\nabla_{\x} \log D_{\phi}(R_{\varphi}(\x))$ is well-defined and computable for both $\x = \x_d$ and $\x = \x_c$. This motivates the ``reward approximation'' approach $\mathbb{E}_{p_(\x_d | \x_c)} \left[ \log D_{\phi}(R_{\varphi}(\x_d)) \right] \approx \log D_{\phi}(R_{\varphi}(\x_c)) $ which sidesteps the intractability of $S_G(\x_c)$ via the reparameterisation trick:
\begin{equation}
\hspace{-0.7em}
\resizebox{0.92\hsize}{!}{$
\begin{aligned}
    \nabla_{\theta} \mathcal{L}_{\text{gen}}(\theta) 
    &\approx \mathbb{E}_{p_G(\x_c)} \left[ \mathbb{E}_{p_(\x_d | \x_c)} \left[ \log D_{\phi}(R_{\varphi}(\x_d)) \right] S_G(\x_c) \right] \\
    &\approx \mathbb{E}_{p_G(\x_c)} \left[ \log D_{\phi}(R_{\varphi}(\x_c)) \nabla_{\theta} \log p_G(\x_c) \right]\\
    &= \mathbb{E}_{p(\z)} \left[ \nabla_{\theta} G_{\theta}(\z) \nabla_{\x_c} \log D_{\phi}(R_{\varphi}(\x_c))|_{\x_c = G_{\theta}(\z)} \right] \\
    &= \nabla_{\theta} \mathbb{E}_{p_G(\x_c)} \left[ \log D_{\phi}(R_{\varphi}(\x_c)) \right] := \nabla_{\theta} \tilde{\mathcal{L}}_{\text{gen}}(\theta).
\end{aligned}$
}
\label{eq:reinforce_gradient_approx}
\end{equation}
The reward approximation quality is key to the performance of generative model, as gradient ascent using (\ref{eq:reinforce_gradient_approx}) would encourage the generator to create continuous voxel grids $\x_c$ for which the neural renderer would return realistic rendering results. Notice the neural renderer is free to render arbitrary outcomes for continuous voxel grids if it is only trained by minimising $\mathcal{L}_{2}(\varphi)$ on discrete voxel grids. Therefore $\x_c$ is not required to resemble the desired 3D shape in order to produce satisfactory neural rendering results. Since the 3D generative model $G_{\theta}(\cdot)$ typically creates non-discrete voxel grids, the generated 2D images using the off-the-shelf renderer will match poorly to the training images.
We address this issue by training the neural renderer with a novel loss function which we call \emph{discriminator output matching} (DOM). Define $F(\cdot)= \log D_{\phi}(\cdot)$, the DOM loss is
\begin{equation}
\hspace{-0.7em}
\resizebox{0.92\hsize}{!}{
    $\mathcal{L}_{\text{DOM}}(\varphi) = \mathbb{E}_{p_G(\x_c)p(\x_d | \x_c)} \left[ ( F(R_d(\x_d)) - F(R_{\varphi}(\x_c)) )^2 \right] .$
}
\end{equation}
Using neural networks of enough capacity, the optimal neural renderer achieves perfect reward approximation, i.e.~$\log D_{\phi}(R_{\varphi^*}(\x_c)) = \mathbb{E}_{p_(\x_d | \x_c)} \left[ \log D_{\phi}(R_d(\x_d)) \right]$.
This approach forces the neural renderer  to preserve the population statistics of the discrete rendered images defined by the discriminator. Therefore to fool the discriminator, the 3D generative model must generate continuous voxel grids which correspond to meaningful representations of the underlying 3D shapes. In practice the neural renderer is trained using a combination of the two loss functions 
\begin{equation}
\min_{\varphi} \mathcal{L}_{\text{render}}(\varphi) := \mathcal{L}_{2}(\varphi) + \lambda \mathcal{L}_{\text{DOM}}(\varphi).
\label{eq:RenderNetLoss}
\end{equation}
We name the proposed proposed method \emph{inverse graphics GAN} (IG-GAN), as the neural renderer in back-propagation time ``inverts'' the off-the-shelf renderer and provide useful gradients for the 3D generative model training. The model is also visualised in Figure \ref{fig:method_visualisation}, which is trained in an end-to-end fashion. To speed up the generative model training, the neural renderer can be pretrained on a generic data set, like tables or cubes, that can differ significantly from the 2D images that the generative model is eventually trained on.
\section{Experimental Setup}

\begin{table*}
\caption{\label{tbl:maintable} FID scores computed on ShapeNet objects (bathtubs, couches and chairs).
\vspace{-1em}}
\begin{center}
 \begin{tabular}{c|ccc|ccc|cccc}
 \Xhline{3\arrayrulewidth}
 \# of Images&\multicolumn{3}{c|}{\bf $500$}&\multicolumn{3}{c|}{\bf One Per Model ($\approx 3000$)}&\multicolumn{3}{c}{\bf Unlimited}\\
 Dataset &{\bf Tubs}&{\bf Couches}&{\bf Chairs}&{\bf Tubs}&{\bf Couches}&{\bf Chairs}&{\bf Tubs}&{\bf Couches}&{\bf Chairs}&{\bf LVP}\\
 \Xhline{2\arrayrulewidth}
 2D-DCGAN&$737.7$&$540.5$&$672.8$&$461.8$&$354.3$&$362.3$&$226.7$&$210.9$&$133.2$&$237.9$\footnotemark[2]\\
\hline
 Visual Hull&$305.8$&$279.3$&$183.4$&$184.6$&$106.2$&$37.1$&$90.1$&$35.1$&$15.7$&$34.5$\\
\hline
 Absorbtion Only&$336.9$&$282.9$&$218.2$&$275.8$&$78.0$&$32.8$&$104.5$&$25.5$&$23.8$&$38.6$\\
 \Xhline{2\arrayrulewidth}
 IG-GAN (Ours)&${\bf187.8}$&${\bf114.1}$&${\bf119.9}$&${\bf67.5}$&${\bf35.8}$&${\bf20.7}$&${\bf44.0}$&${\bf17.8}$&${\bf13.6}$&${\bf20.6}$\\
\Xhline{3\arrayrulewidth}
 \end{tabular}
\end{center}
\end{table*}

\subsection{Implementation Details}

\paragraph{Off-the-shelf Renderer}
Our rendering engine is based on the Pyrender~\cite{pyrender} which is built on top of OpenGL. 

\paragraph{Architecture}
We employ a 3D convolutional GAN architecture for the generator ~\cite{wu2016learning} with a $64^3$ voxel resolution. To incorporate the viewpoint, the rigid body transformation embeds the $64^3$ grid into a $128^3$ resolution. We render the volumes with a RenderNet ~\cite{rendernet} architecture, with the modification of using 2 residual blocks in 3D and 4 residual blocks in 2D only. The discriminator architecture follows the design in DCGAN \cite{DCGAN}, taking images of $128^2$ resolution. Additionally, we add spectral normalization to the discriminator \cite{specNorm} to stablize training.

\paragraph{Hyperparameters}
We employ a 1:1:1 updating scheme for generator, discriminator and the neural renderer, using learning rates of $2\mathrm{e}{-5}$ for the neural renderer and $2\mathrm{e}{-4}$ for both generator and discriminator.  The Discriminator Output Matching loss is weighted by $\lambda = 100$ over the $\mathcal{L}_2$ loss, as in \eqref{eq:RenderNetLoss}. We found that training was stable against changes in $\lambda$ and extensive tuning was not necessary.  The binerization distribution $p(\x_d | \x_c)$ was chosen as a global thresholding, with the threshold being distributed uniformly in $[0,1]$.

\subsection{Datasets}
We evaluate our model on a variety of synthetic datasets generated from 3D models in the ShapeNet ~\cite{ShapeNet} database as well as on a real data set consisting of chanterelle mushrooms, introduced in ~\cite{platosCave}.

We synthesize images from three different categories of ShapeNet objects, \textit{Chairs, Couches} and \textit{Bathtubs}. For each of these categories, we generate three different datasets:

\noindent\textbf{500:} A very small dataset consisting of 500 images taken from 500 different objects in the corresponding data set, each rendered from a single viewpoint. 

\noindent\textbf{One per Model:}  A more extensive one where we sample each object in the 3D data set once from a single viewpoint for the \textit{Chairs} and \textit{Couches} data, and from four viewpoints for \textit{Bathtubs}, due to the small number of objects in this category. This results in data sets containing 6777 chairs, 3173 couches and 3424 images of bathtubs. 

\noindent\textbf{Unlimited:} Finally, we render each objects from a different viewpoint throughout each training epoch. This leads to a theoretically unlimited data set of images. However, all images are generated from the limited number of objects contained in the corresponding ShapeNet category.

In order to render the ShapeNet volumes, we light them from two beam light sources that illuminate the object from 45$^{\circ}$ below the camera and 45$^{\circ}$ to its left and right.  The camera always points directly at the object, and we uniformly sample from all 360$^{\circ}$ of rotation, with an elevation between 15 and 60 degrees for bathtubs and couches and -30 and 60 degrees for chairs. The limitation on elevation angle is chosen to generate a dataset that is as realistic as possible, given the object category.  

\noindent\textbf{LVP:}  We also consider a limited viewpoint (LVP) setting for the  chairs dataset where the azimuth rotation is limited to 60$^{\circ}$ in each direction from center, and elevation is limited to be between 15 ad 60 degrees.  This setting is meant to further simulate the viewpoint bias observed in natural photographs.

\noindent\textbf{Chanterelle Mushrooms:} We prepare the \textit{Chanterelle} data by cropping and resizing the images to $128^2$ resolution and by unifying the mean intensity in an additive way. Note that the background of the natural images has been masked out in the original open-sourced dataset.

\subsection{Baselines}

We compare to the following state-of-the-art methods for learning 3D voxel generative models from 2D data:

\noindent\textbf{Visual Hull:} This is the model from \citet{PrGAN} which learned from a smoothed version of object silhouettes.  

\noindent\textbf{Absorbtion Only:}  This is the model from \citet{platosCave} which assumes voxels absorb light based on their fraction of occupancy.

\noindent\textbf{2D-DCGAN:}  For comparison we also show results from a DCGAN~\cite{DCGAN} trained on 2D images only. While this baseline is solving a different task and is not able to produce any 3D objects, it allows a comparison of the generated image quality.

We show results from our reimplementation of these models because we found~\citet{PrGAN} performed better by using recently developed GAN stabalization techniques, i.e. spectral normalization~\cite{specNorm}, and the code for~\citet{platosCave} was not available at the time we ran our original results. A discussion of the Emission-Absorption model also proposed in ~\citet{platosCave} is provided in the supplemental material \footnote{\url{https://lunz-s.github.io/iggan/iggan_supplemental.pdf}}. To provide the most favorable comparison for the baselines, each baseline is trained on a dataset of images synthesized from ShapeNet using the respective choice of rendering (Visual Hull or Absorbtion-Only) from the 3D objects.


\subsection{Evaluation Metrics}

We chose to evaluate the quality of the generated 3D models by rendering them to 2D images and computing Fréchet Inception Distances (FIDs)~\cite{FID}.
This focuses the evaluation on the observed visual quality, preventing if from considering what the model generates in the unobserved insides of the objects.  All FID scores reported in the main paper use an Inception network~\cite{szegedy2016rethinking} retrained to classify Images generated with our renderer because we found this to better align with our sense of the visual quality given the domain gap between ImageNet and our setting. We have trained the Inception network to classify rendered images of the 21 largest object classes in ShapeNet, achieving 95.3\% accuracy. In the supplemental material we show FID scores using the traditional Inception network trained on ImageNet~\cite{ImageNet}, and qualitatively they are very similar.


\section{Results}



\subsection{Quantitative Evaluation} 
We can see clearly from Table~\ref{tbl:maintable} that our approach (IG-GAN) significantly out-performs the baselines on all dasasets.  The largest of these gains is obtained on the data sets containing many concavities, like couches and bathtubs. Furthermore, the advantage of the proposed method becomes more significant when the dataset is smaller and the viewpoint is more restrictive.  Since our method can more easily take advantage of the lighting and shading cue provided by the images, we believe it can extract more meaningful information per training sample, hence producing better results in these settings. On the Unlimited dataset the baseline methods seem to be able to mitigate some of their disadvantage by simply seeing enough views of each training model, but still our approach generates considerably better FID scores even in this setting.  We note that the very poor results from the 2D-DCGAN stem from computing FIDs using our retrained Inception network, which seems to easily pick up on any unrealistic artifacts generated by the GAN. The supplemental materials show the FID scores using the standard Inception net, and while IG-GAN still outperforms the 2D-DCGAN considerably, the resulting scores are closer.

\footnotetext[2]{A fair comparison to DCGAN is impossible, as the generator is trained on LVP data, but FID evaluations use a 360 $^{\circ}$ view.}



\begin{figure*}[!htb]
    \centering
    \begin{minipage}{.5\textwidth}
    \includegraphics[width=\textwidth]{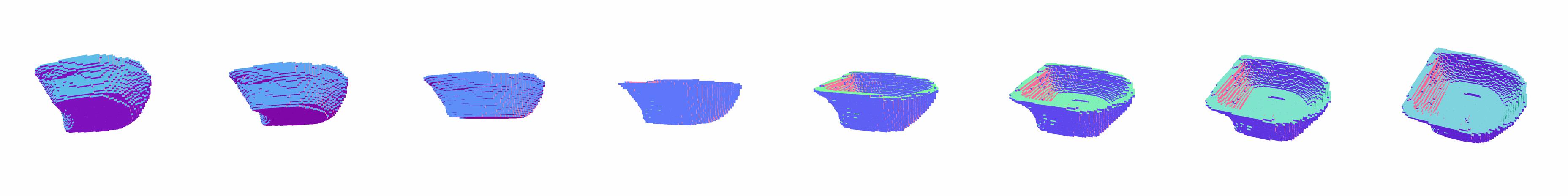}
    \includegraphics[width=\textwidth]{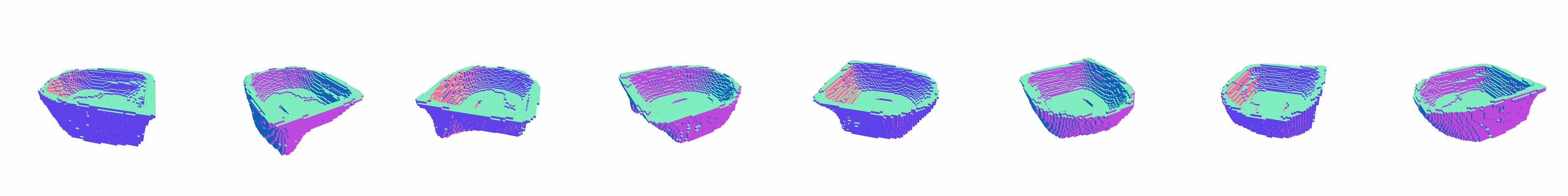}
    \\
    \includegraphics[width=\textwidth]{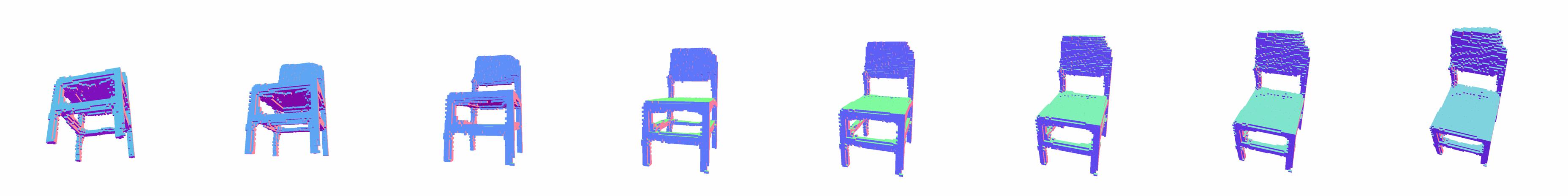}
    \includegraphics[width=\textwidth]{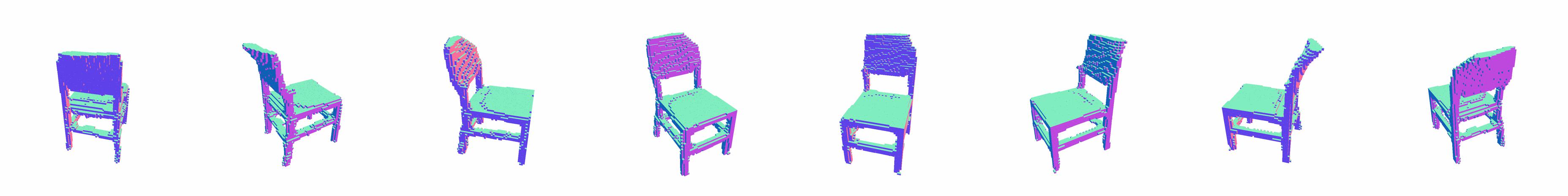}
    \\
    \includegraphics[width=\textwidth]{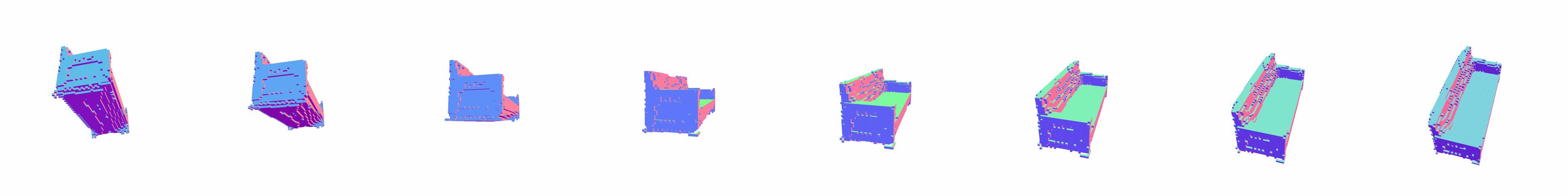}
    \includegraphics[width=\textwidth]{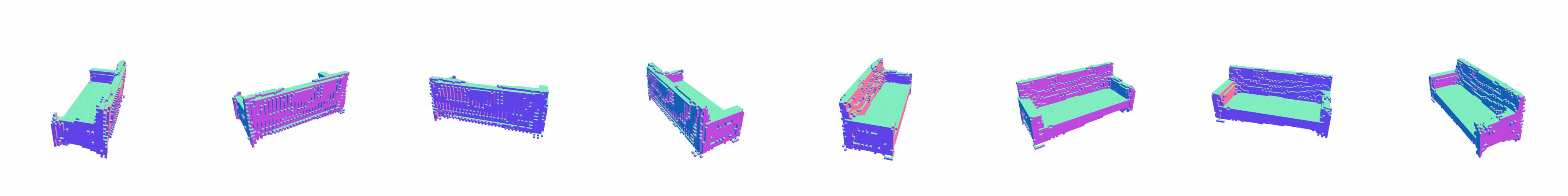}
    \end{minipage}%
    \hfill
    \begin{minipage}{0.45\textwidth}
    \centering
   \includegraphics[width=0.32\textwidth]{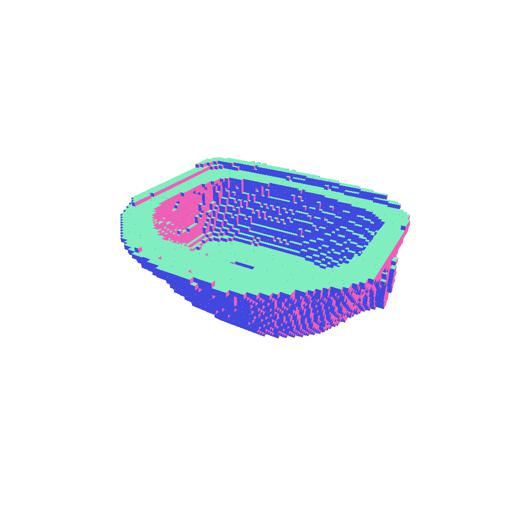}
    \includegraphics[width=0.3\textwidth]{Images/Chairs/Unlimited/REINFORCE/individual_images/00077_th_02_NM.jpg}
    \includegraphics[width=0.3\textwidth]{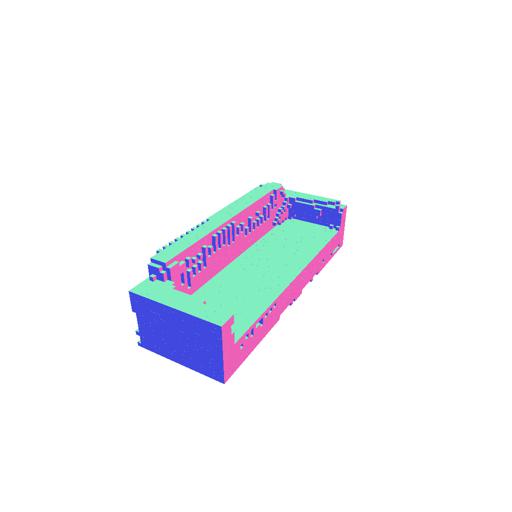}
    \\
    \includegraphics[width=0.32\textwidth]{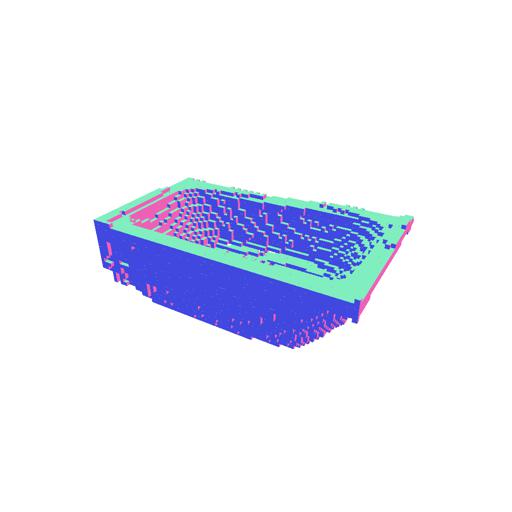}
    \includegraphics[width=0.32\textwidth]{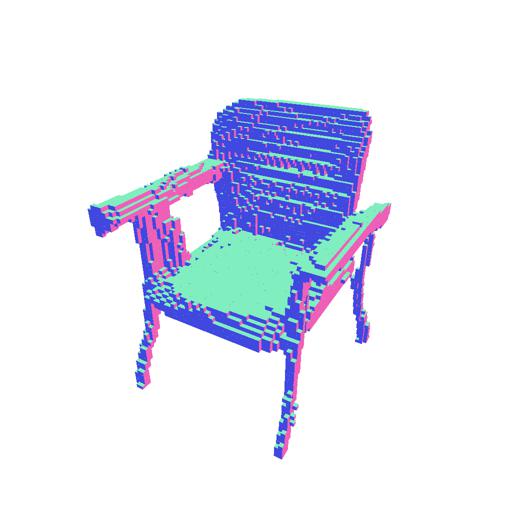}
    \includegraphics[width=0.32\textwidth]{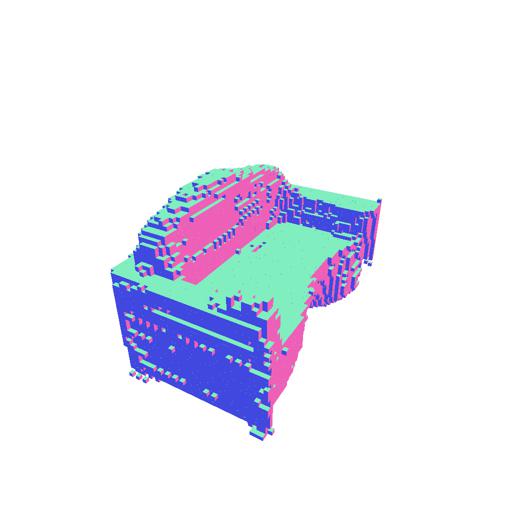}
    \\
    \includegraphics[width=0.32\textwidth]{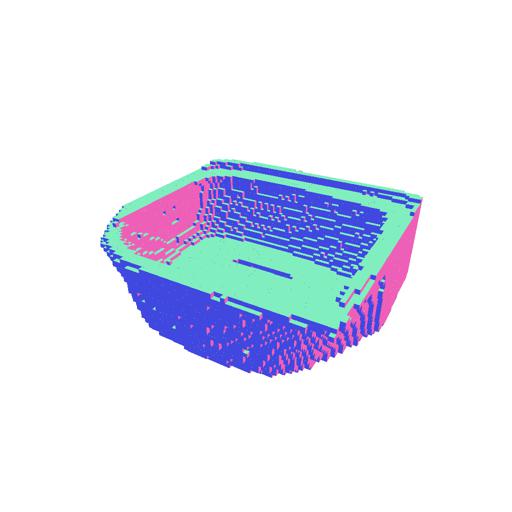}
    \includegraphics[width=0.32\textwidth]{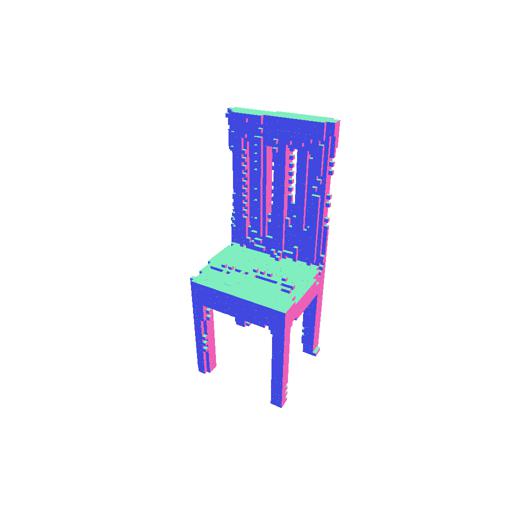}
    \includegraphics[width=0.32\textwidth]{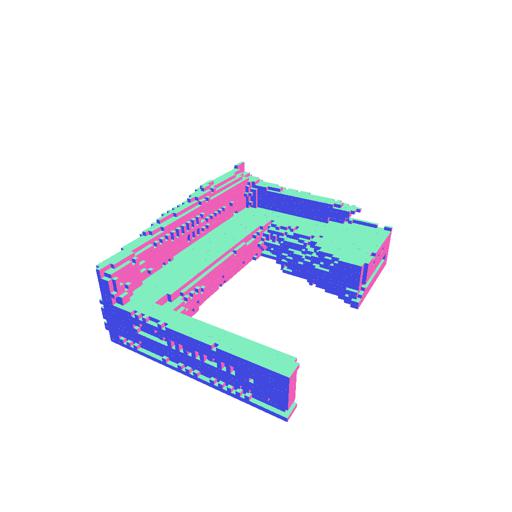}
    \end{minipage}
    
    \caption{\label{fig:IGGANSamples} Normal Maps of objects generated by IG-GAN on the 'Unlimited' datasets. The left panel shows a single sample rendered in different view points, and the right panel shows multiple samples rendered from a canonical viewpoint.}
\end{figure*}

\subsection{Qualitative Evaluation} 
\begin{figure*}
\centering
\begin{subfigure}[t]{.31\textwidth}
    \includegraphics[width=.22\textwidth]{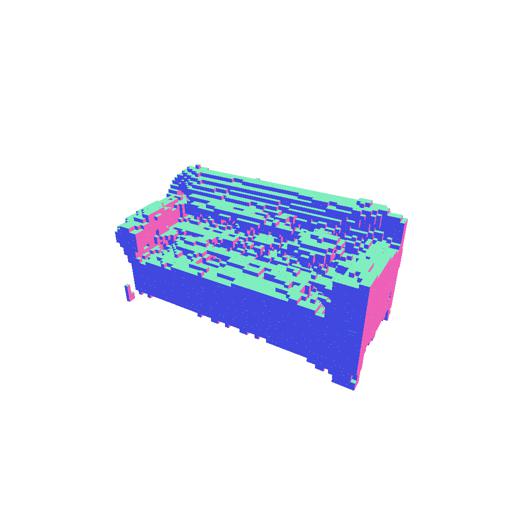}
    \includegraphics[width=.22\textwidth]{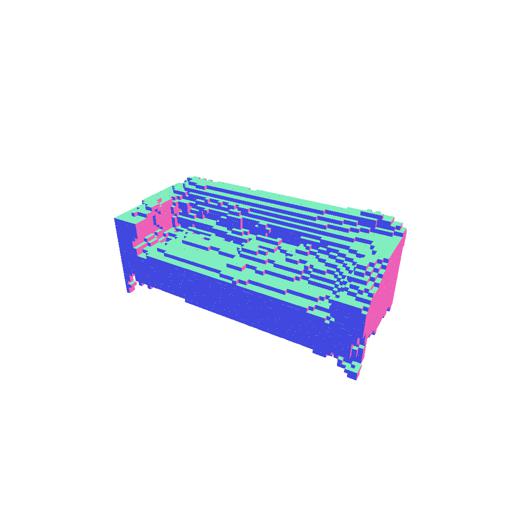}
    \includegraphics[width=.22\textwidth]{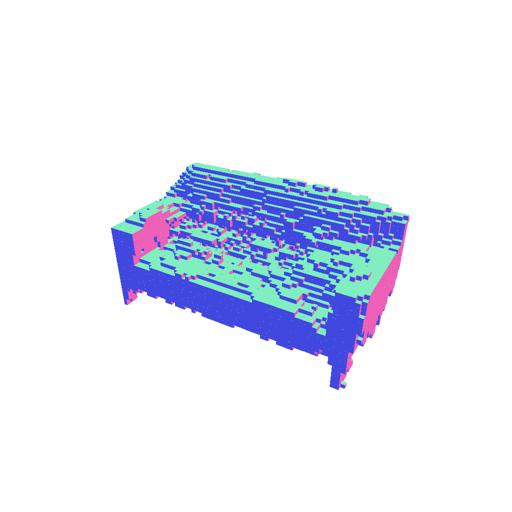}
    \includegraphics[width=.22\textwidth]{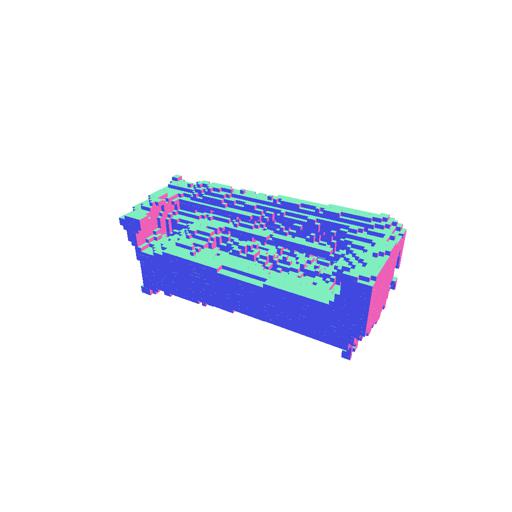}
    \\
    \includegraphics[width=.22\textwidth]{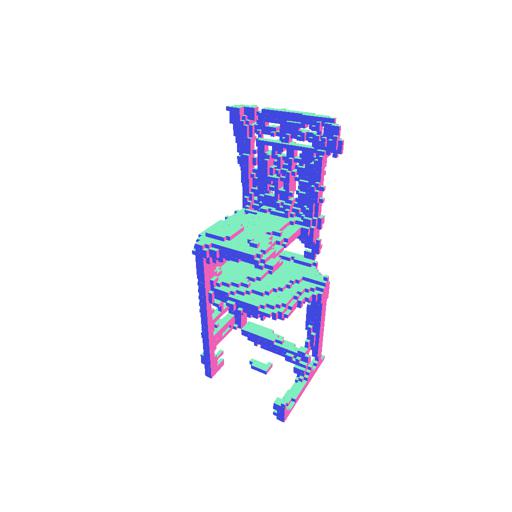}
    \includegraphics[width=.22\textwidth]{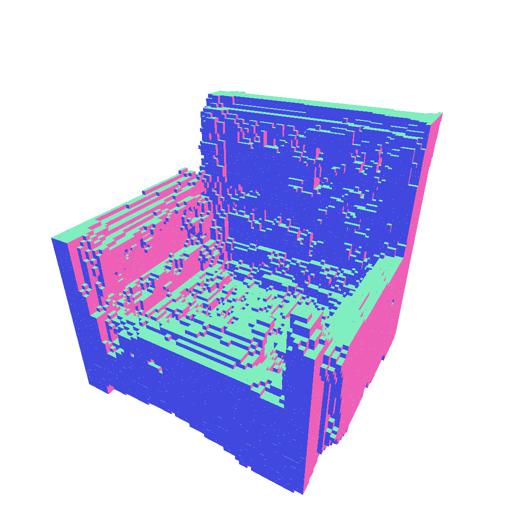}
    \includegraphics[width=.22\textwidth]{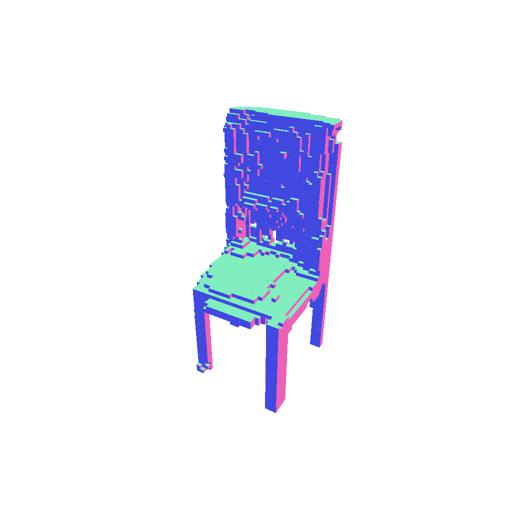}
    \includegraphics[width=.22\textwidth]{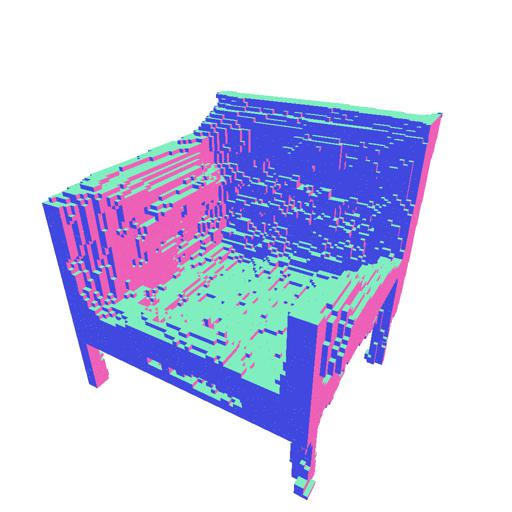}
    \\
    \includegraphics[width=.22\textwidth]{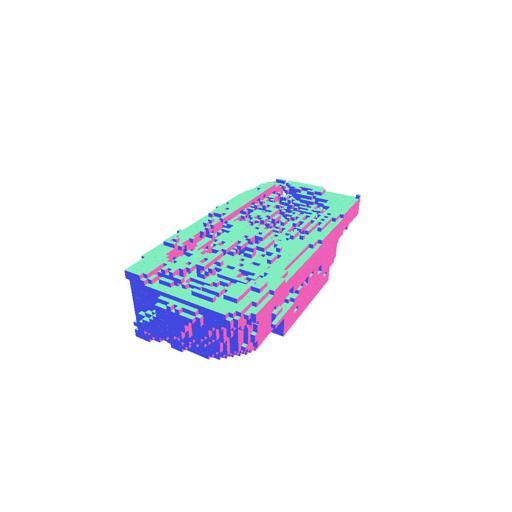}
    \includegraphics[width=.22\textwidth]{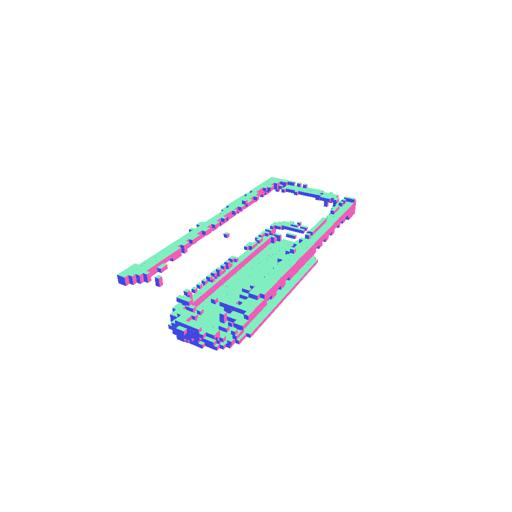}
    \includegraphics[width=.22\textwidth]{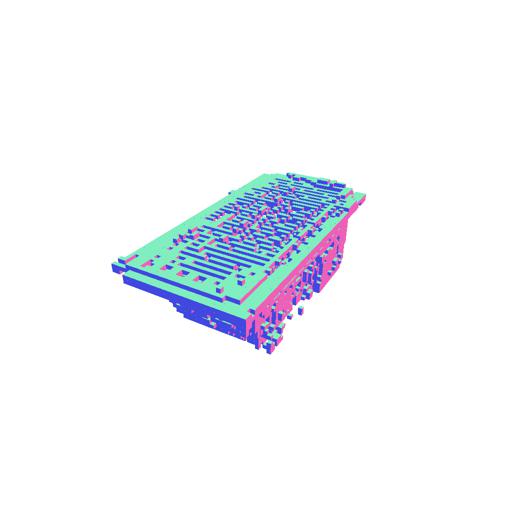}
    \includegraphics[width=.22\textwidth]{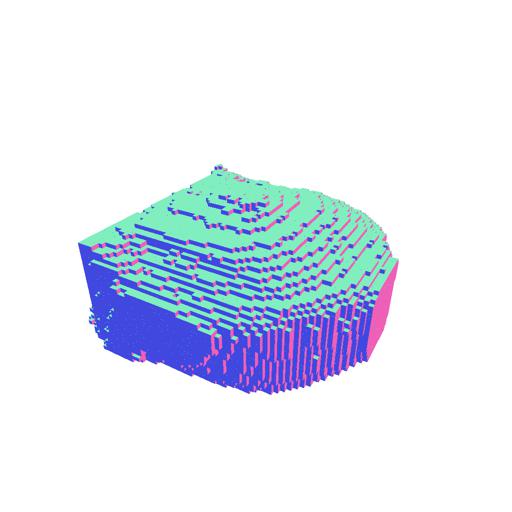}
    \caption{Absorption Only}
\end{subfigure}
\hfill
\begin{subfigure}[t]{.31\textwidth}
    \includegraphics[width=.22\textwidth]{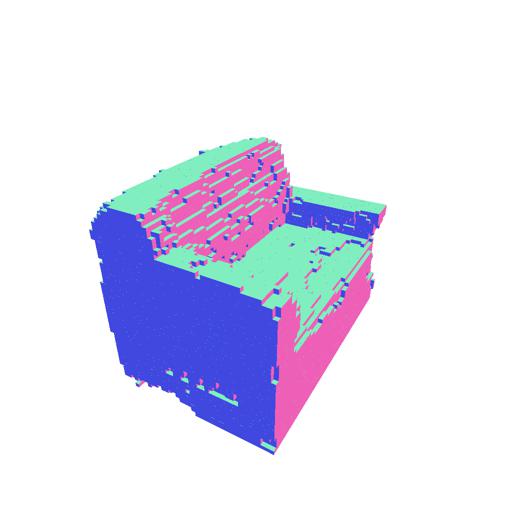}
    \includegraphics[width=.22\textwidth]{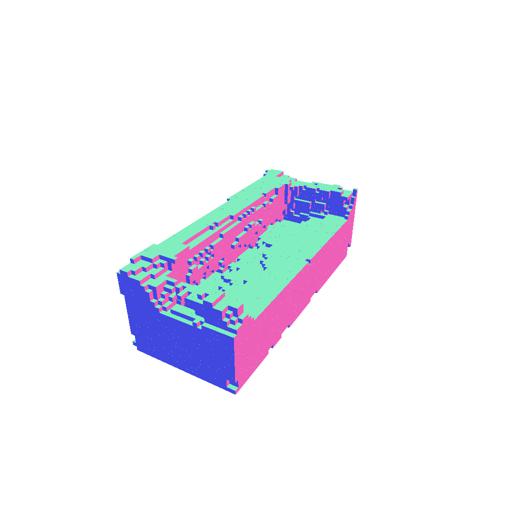}
    \includegraphics[width=.22\textwidth]{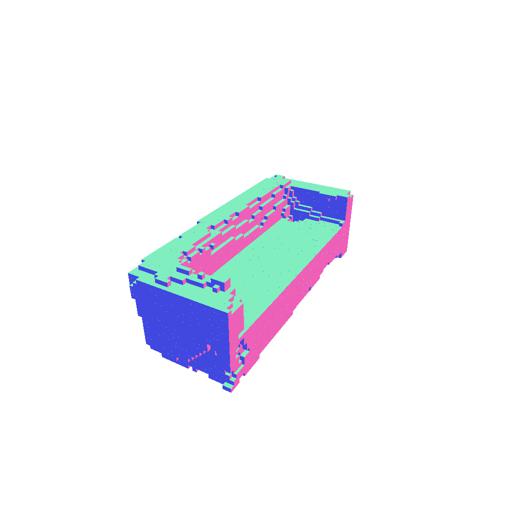}
    \includegraphics[width=.22\textwidth]{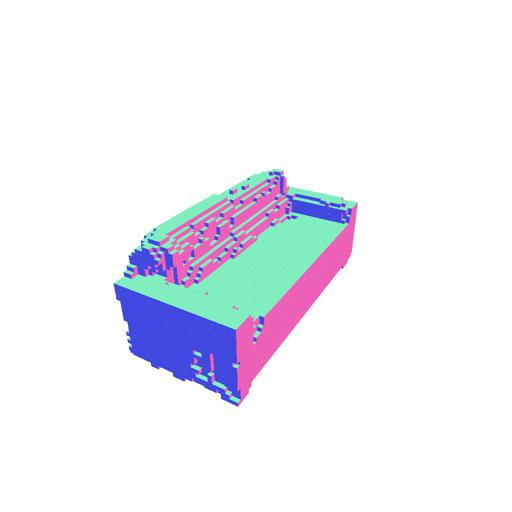}
    \\
    \includegraphics[width=.22\textwidth]{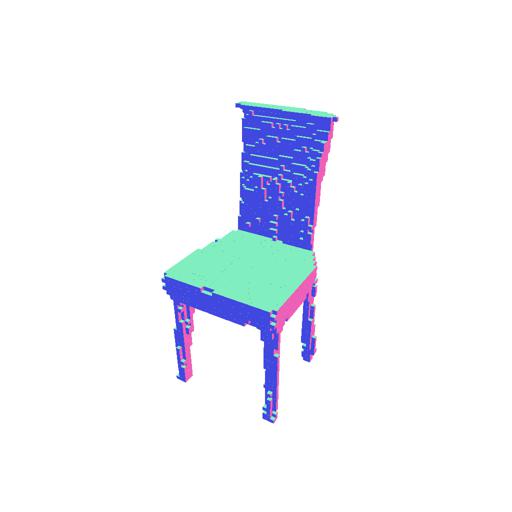}
    \includegraphics[width=.22\textwidth]{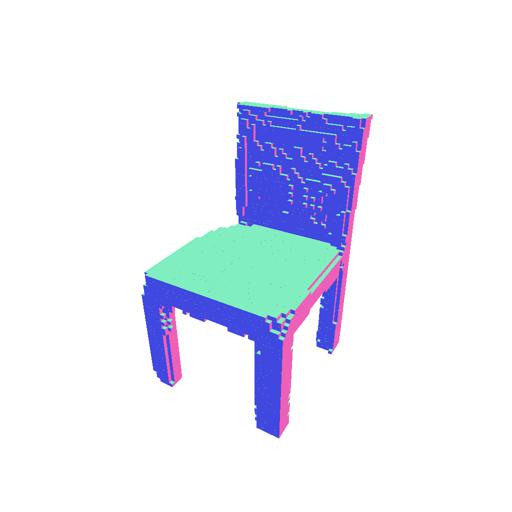}
    \includegraphics[width=.22\textwidth]{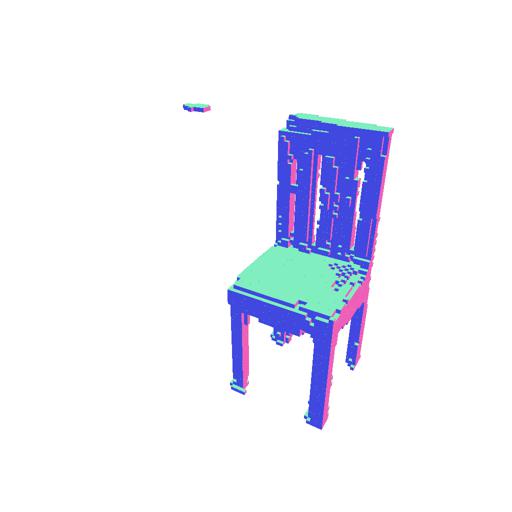}
    \includegraphics[width=.22\textwidth]{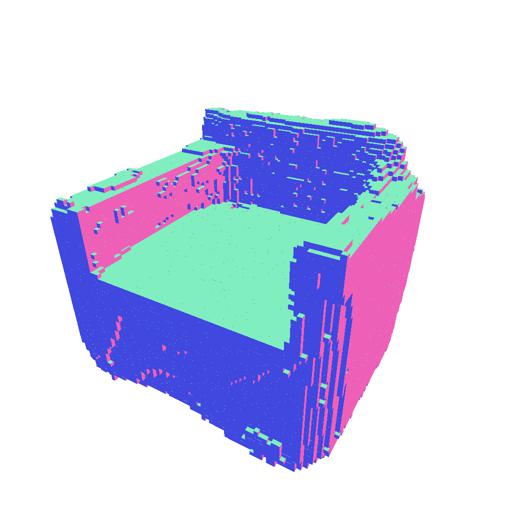}
    \\
    \includegraphics[width=.22\textwidth]{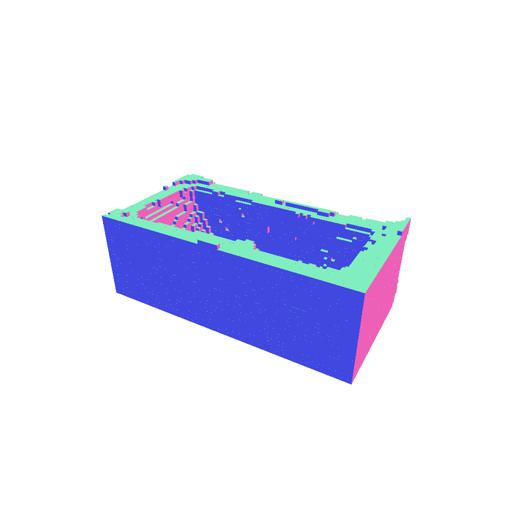}
    \includegraphics[width=.22\textwidth]{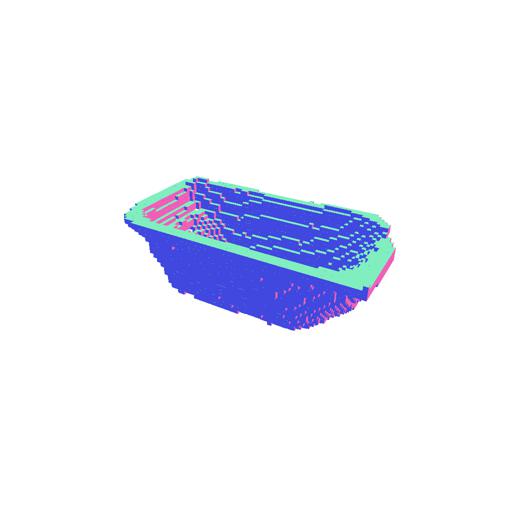}
    \includegraphics[width=.22\textwidth]{Images/Bathtubs/All/REINFORCE/individual_images/00003_th_02_NM.jpg}
    \includegraphics[width=.22\textwidth]{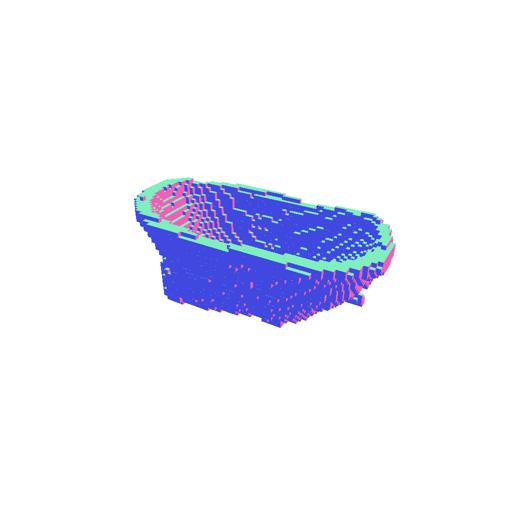}
    \caption{IG-GAN (Ours)}
\end{subfigure}\hfill
\hfill
\begin{subfigure}[t]{.31\textwidth}
    \includegraphics[width=.22\textwidth]{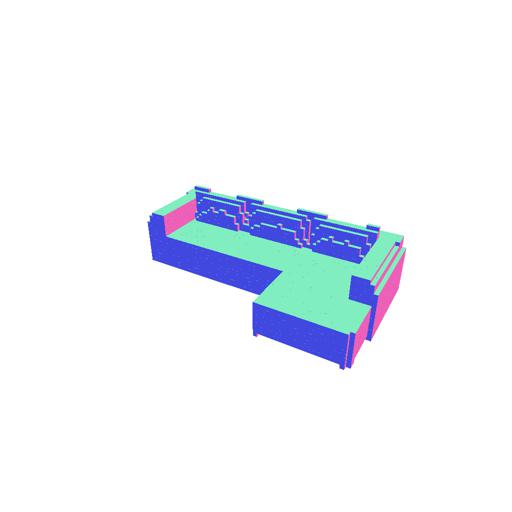}
    \includegraphics[width=.22\textwidth]{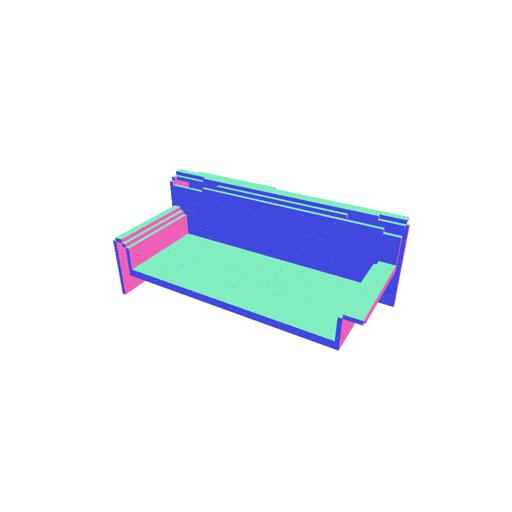}
    \includegraphics[width=.22\textwidth]{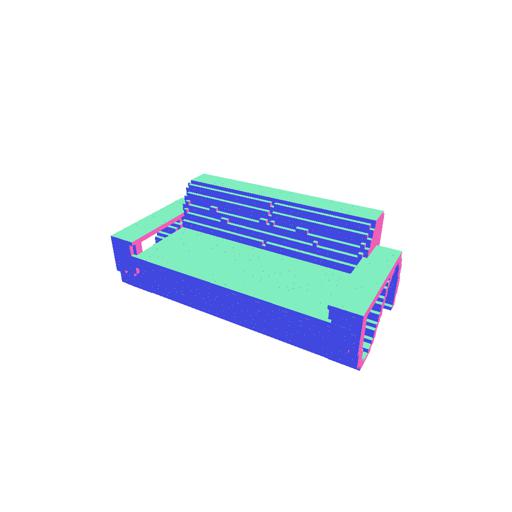}
    \includegraphics[width=.22\textwidth]{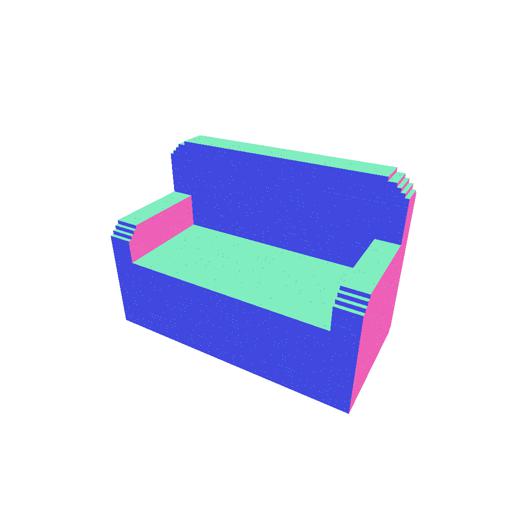}
    \\
    \includegraphics[width=.22\textwidth]{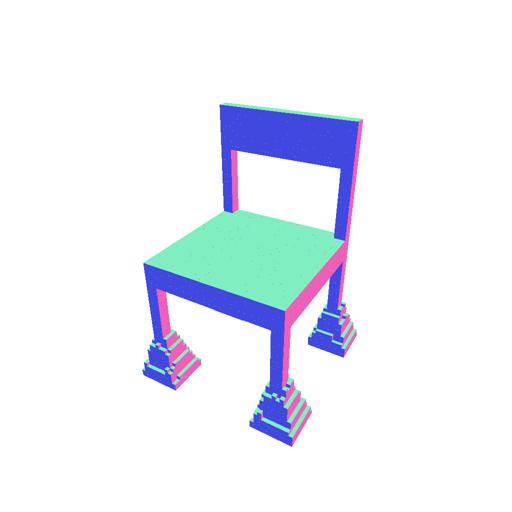}
    \includegraphics[width=.22\textwidth]{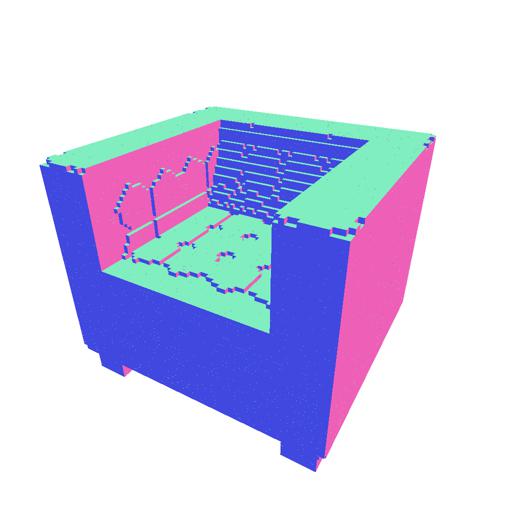}
    \includegraphics[width=.22\textwidth]{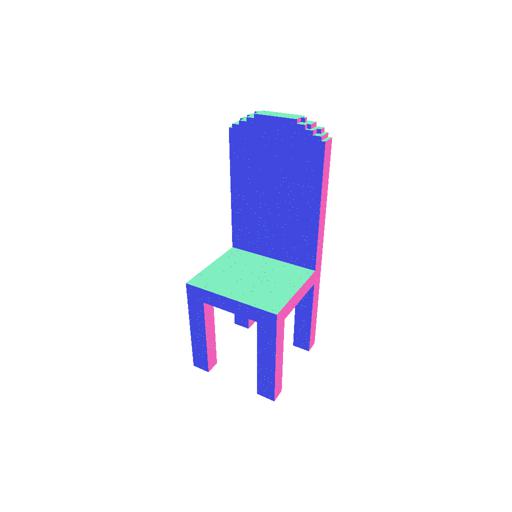}
    \includegraphics[width=.22\textwidth]{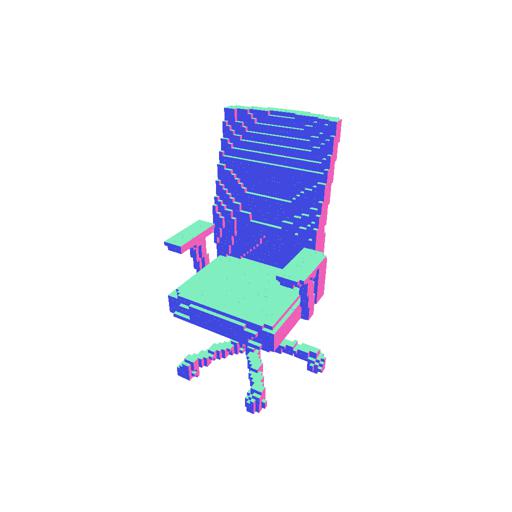}
    \\
     \includegraphics[width=.22\textwidth]{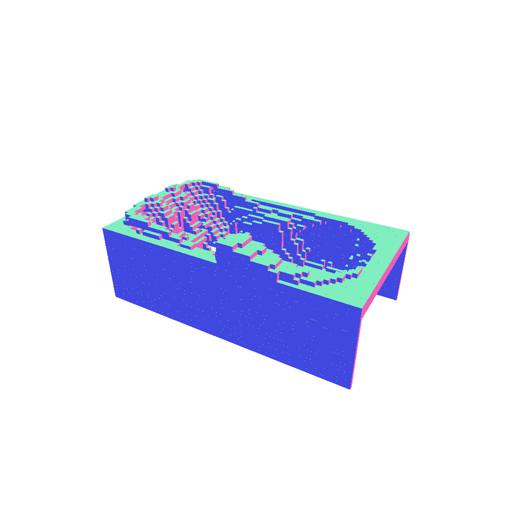}
     \includegraphics[width=.22\textwidth]{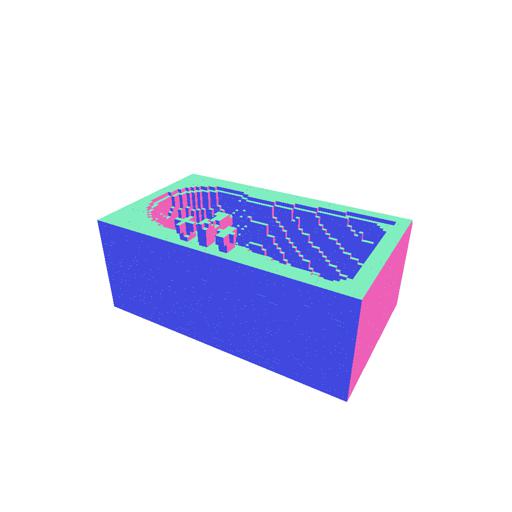}
     \includegraphics[width=.22\textwidth]{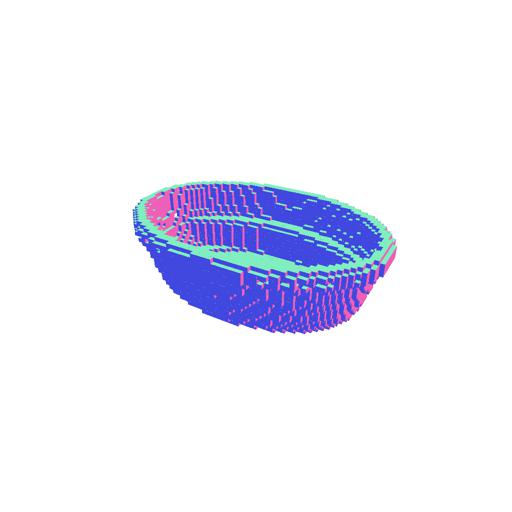}
     \includegraphics[width=.22\textwidth]{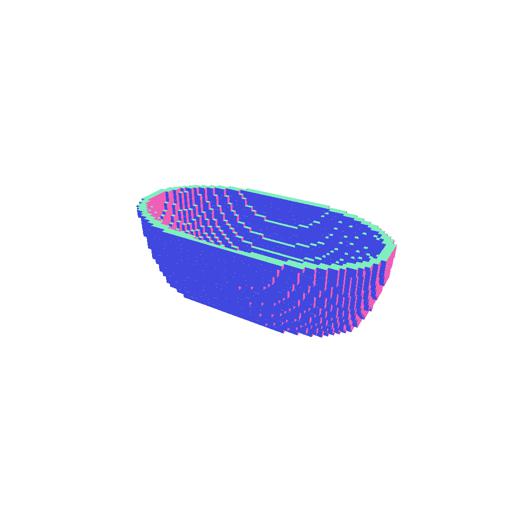}
    \caption{Data Set}
\end{subfigure}

\caption{Samples generated by (a) the AO baseline and (b) our model on the 'One per model' setting (Dataset samples in panel (c)). The samples from the VH baseline is visually similar to those from the AO baseline. Our method is able to recognize concavities correctly, leading to realistic samples of bathtubs and couches.}
\label{fig:Comparison}
\end{figure*}

From Figure \ref{fig:IGGANSamples} we can see that IG-GAN produces high-quality samples on all three object categories. Furthermore,  we can see from Figure \ref{fig:Comparison} that the generated 3D shapes are superior to the baselines.  This is particularly evident in the context of concave objects like bathtubs or couches. Here, generative models based on visual hull or absorption rendering fail to take advantage of the shading cues needed to detect the hollow space inside the object, leading to e.g.~seemingly filled bathtubs. The shortcoming of the baseline models has already been noticed by \citet{PrGAN}. Our approach, on the other hand, successfully detects the interior structure of concave objects using the differences in light exposures between surfaces, enabling it to accurately capture concavities and hollow spaces.

\begin{figure}
    \centering
    \begin{subfigure}[t]{.45\textwidth}
    \includegraphics[width=.19\textwidth]{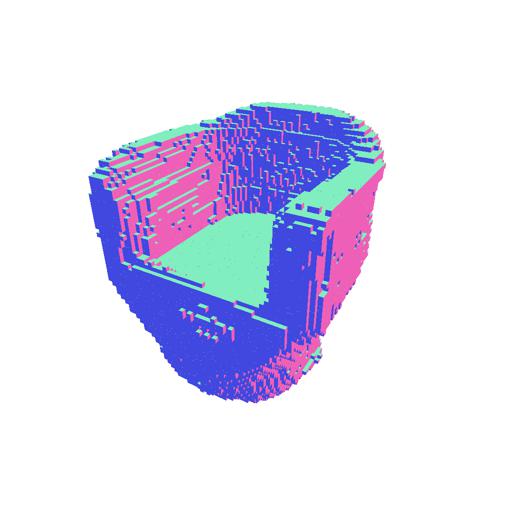}
    \includegraphics[width=.19\textwidth]{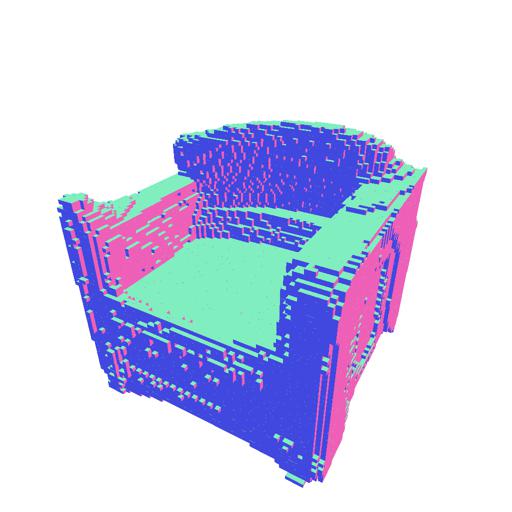}
    \includegraphics[width=.19\textwidth]{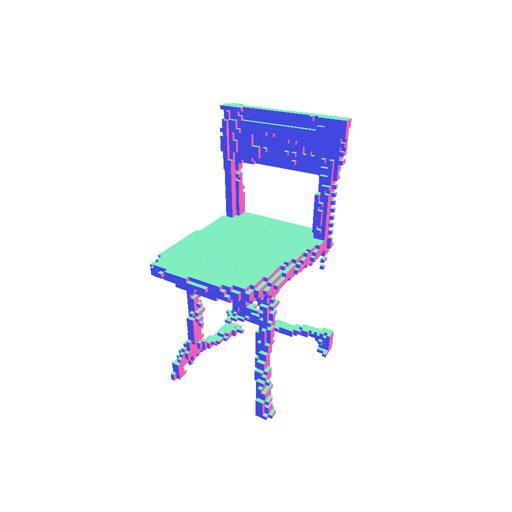}
    \includegraphics[width=.19\textwidth]{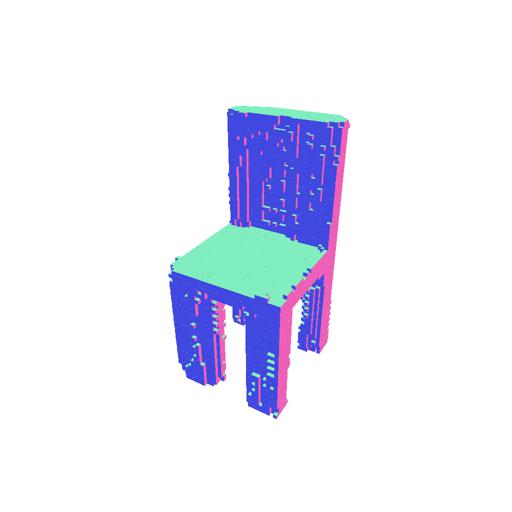}
    \includegraphics[width=.19\textwidth]{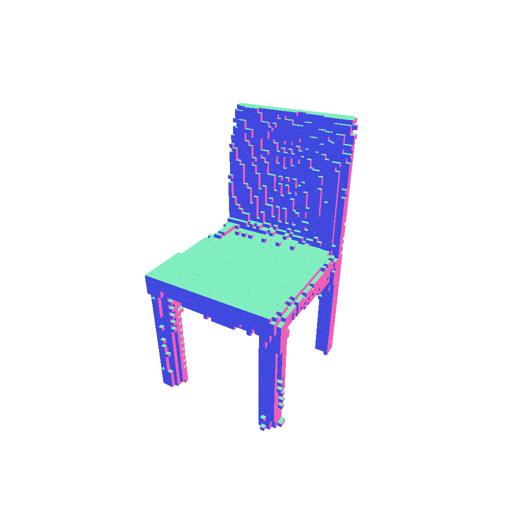}
    \caption{IG-GAN (Ours)}
    \end{subfigure}
    \begin{subfigure}[t]{.45\textwidth}
    \includegraphics[width=.19\textwidth]{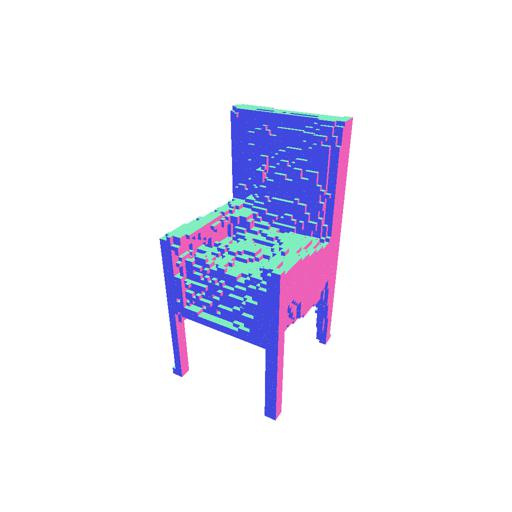}
    \includegraphics[width=.19\textwidth]{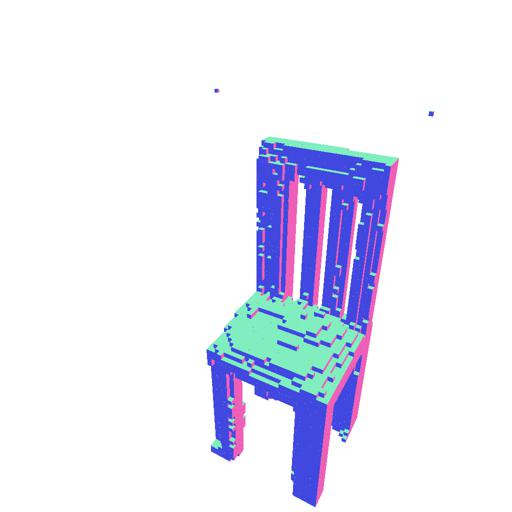}
    \includegraphics[width=.19\textwidth]{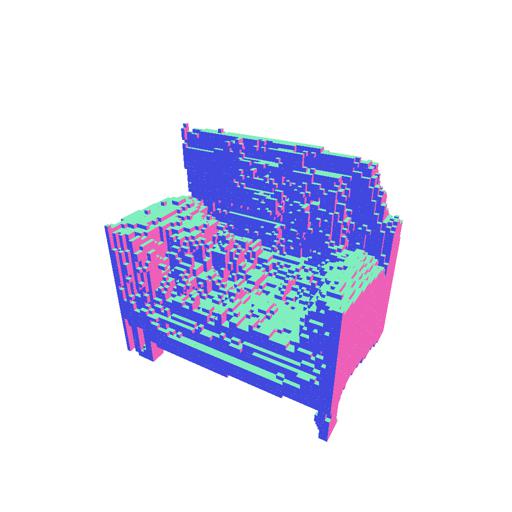}
    \includegraphics[width=.19\textwidth]{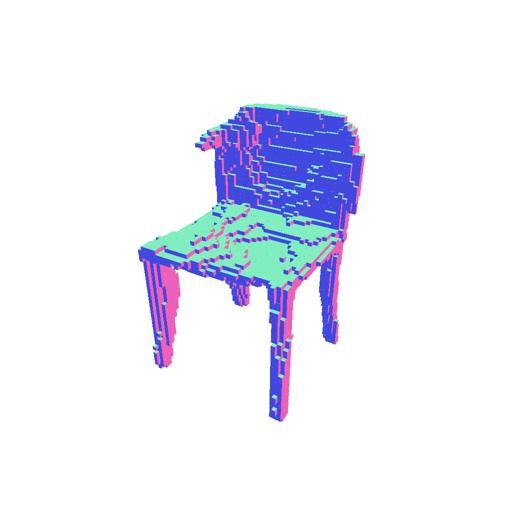}
    \includegraphics[width=.19\textwidth]{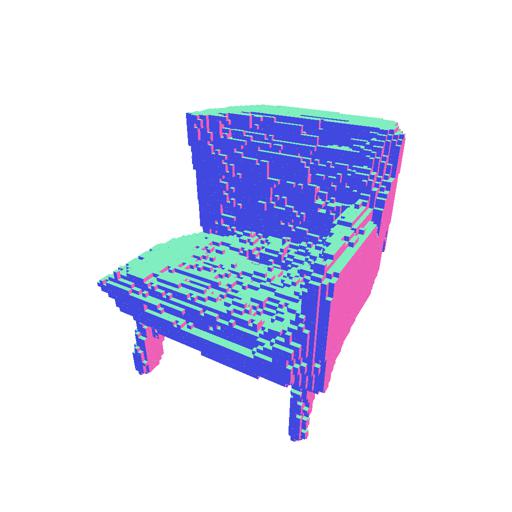}
    \caption{Absorption Only}
    \end{subfigure}
    \caption{Results on the chairs LVP dataset. Unlike our method, the baseline can not extract sufficient information from the data to create chair samples with flat surfaces.}
    \label{fig:ChairsComparisonLVP}
\end{figure}

On the chair dataset, the advantages of our proposed method are evident on flat surfaces. Any uneven surfaces generated by mistake are promptly detected by our discriminator which can easily differences in light exposure, forcing the generator to produce clean and flat surfaces. The baseline methods however are unable to render such impurities in a way that is evident to the discriminator, leading to generated samples with grainy and uneven surfaces. This effect is most obvious on the chairs with limited views (LVP), as shown in Figure \ref{fig:ChairsComparisonLVP}. A large selection of randomly generated samples from all methods can be found in the supplemental material.

\begin{figure}[h]
    \centering
    \includegraphics[width=.15\textwidth]{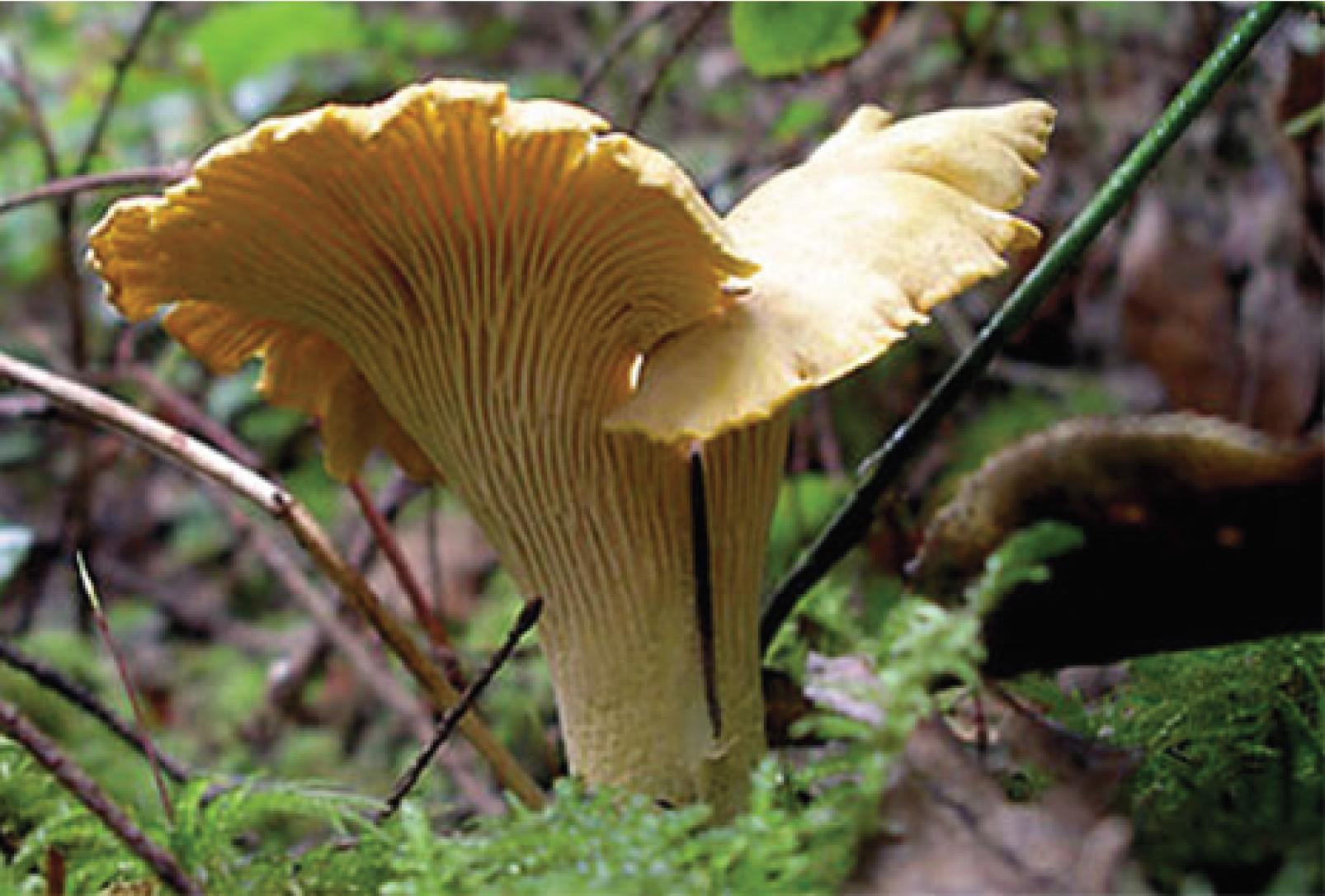}
    \includegraphics[width=.15\textwidth]{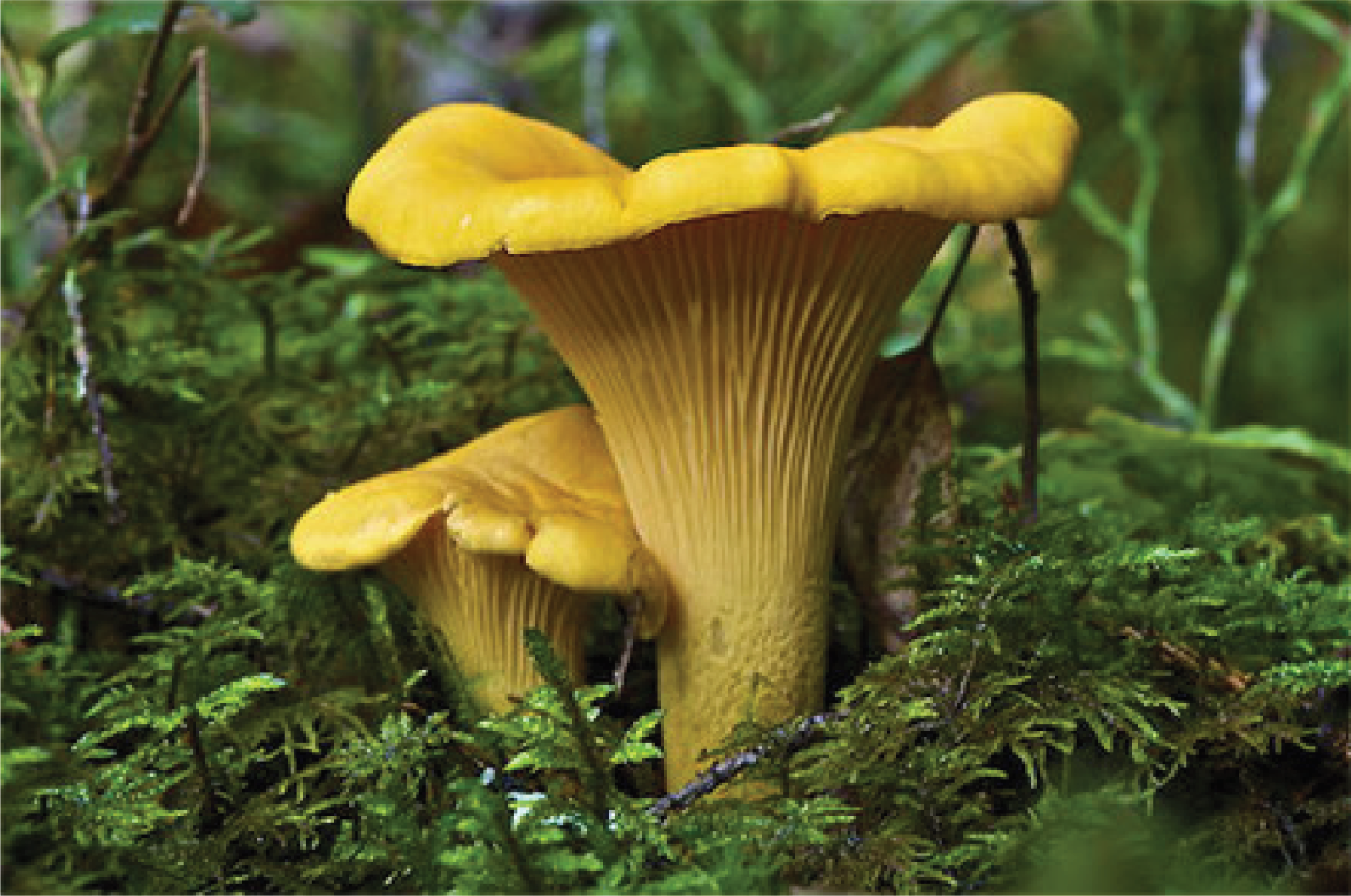}
    \includegraphics[width=.15\textwidth]{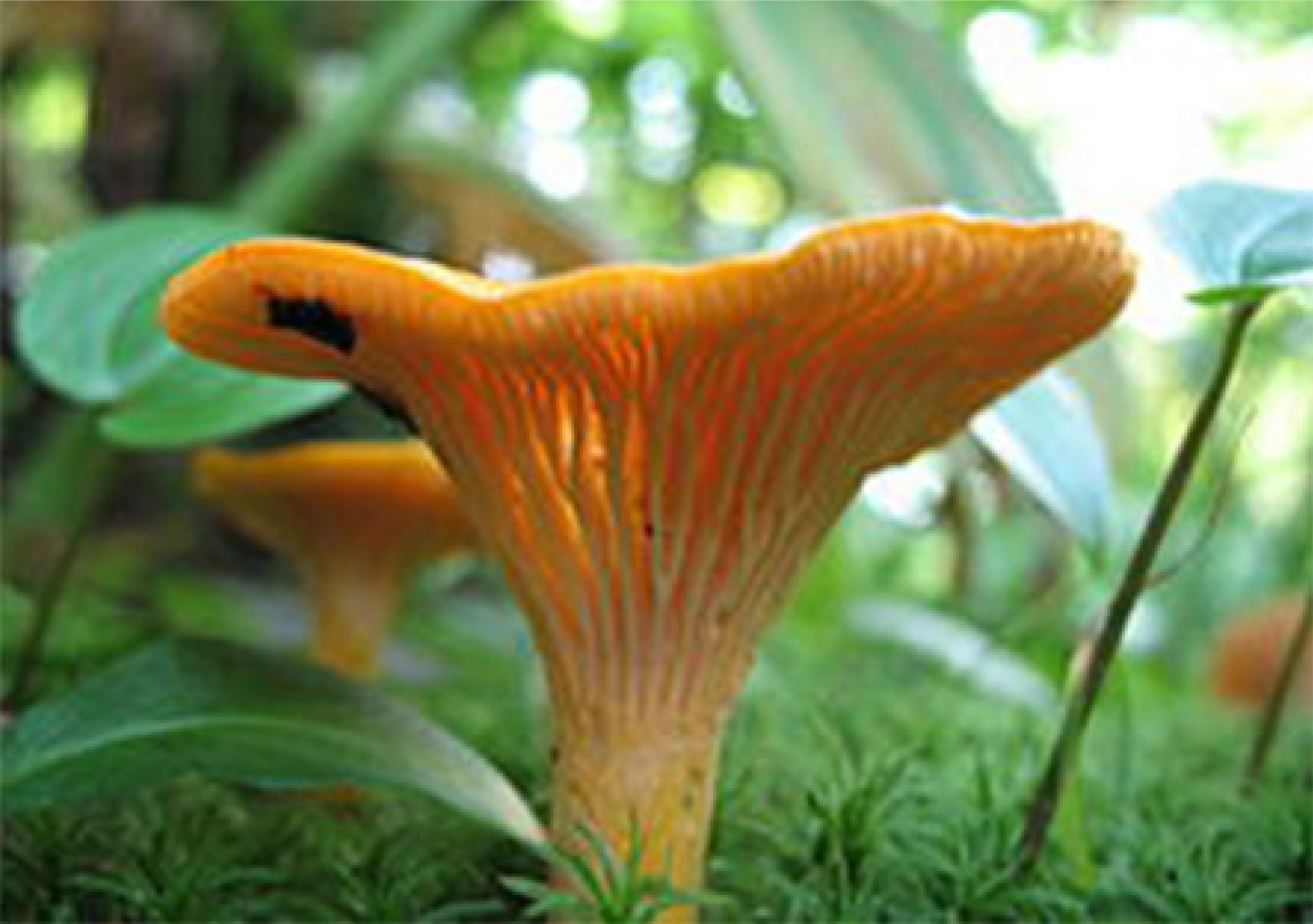}
    \\
    \includegraphics[width=.15\textwidth]{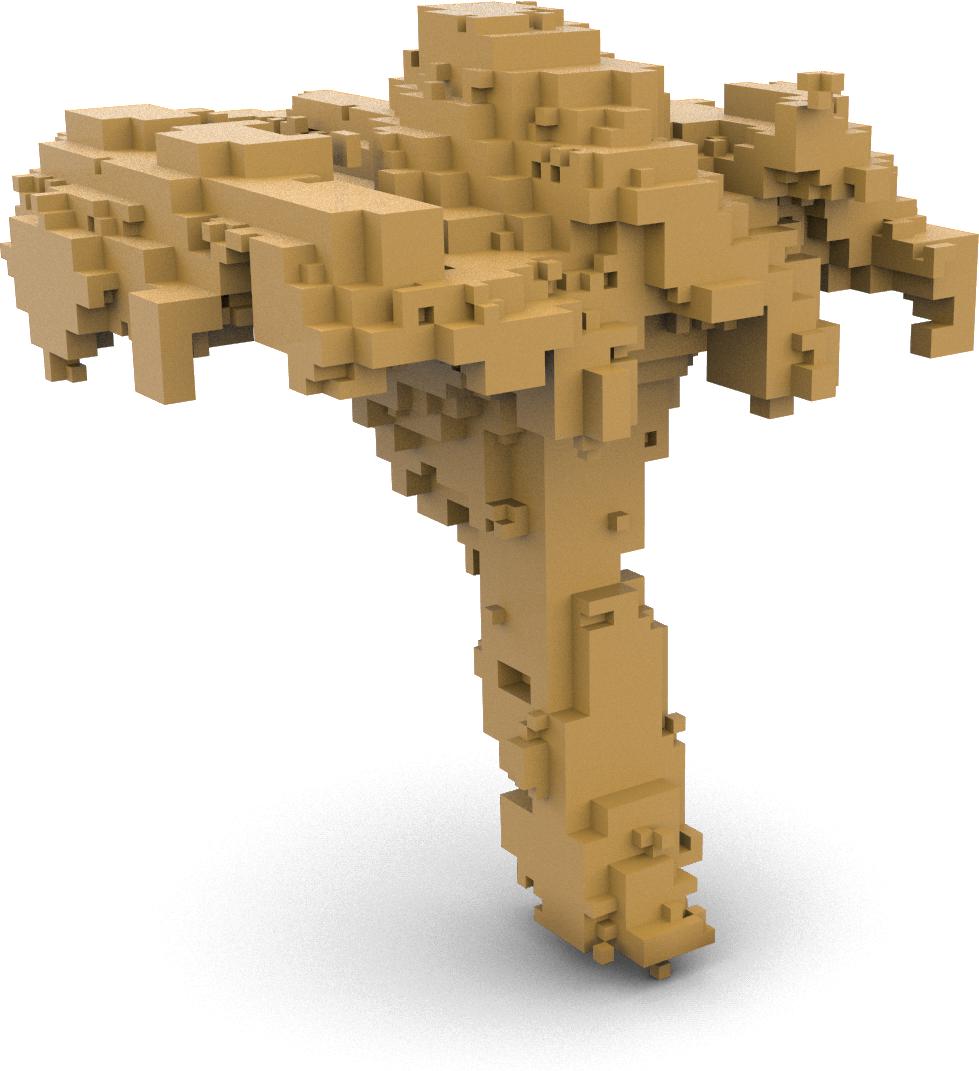}
    \includegraphics[width=.15\textwidth]{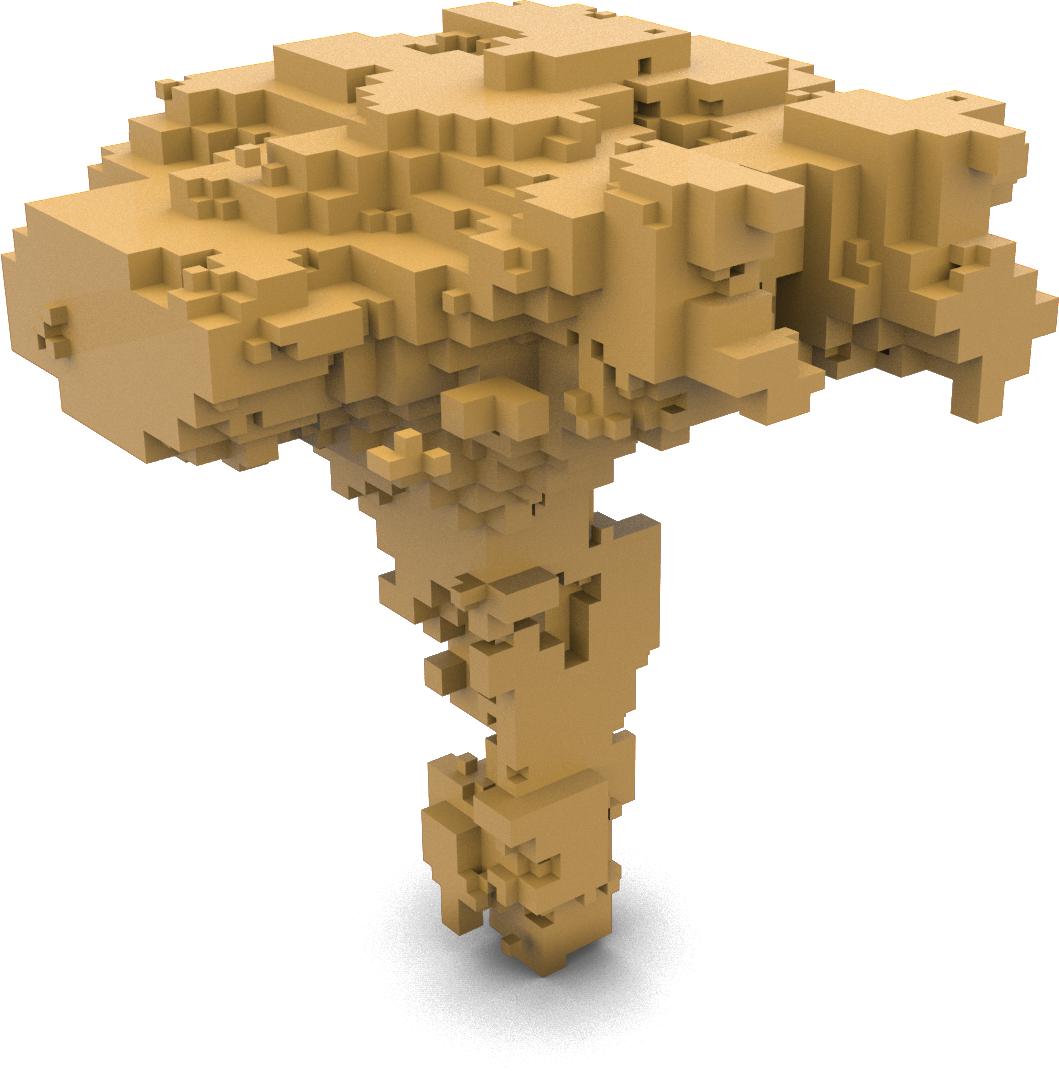}
    \includegraphics[width=.15\textwidth]{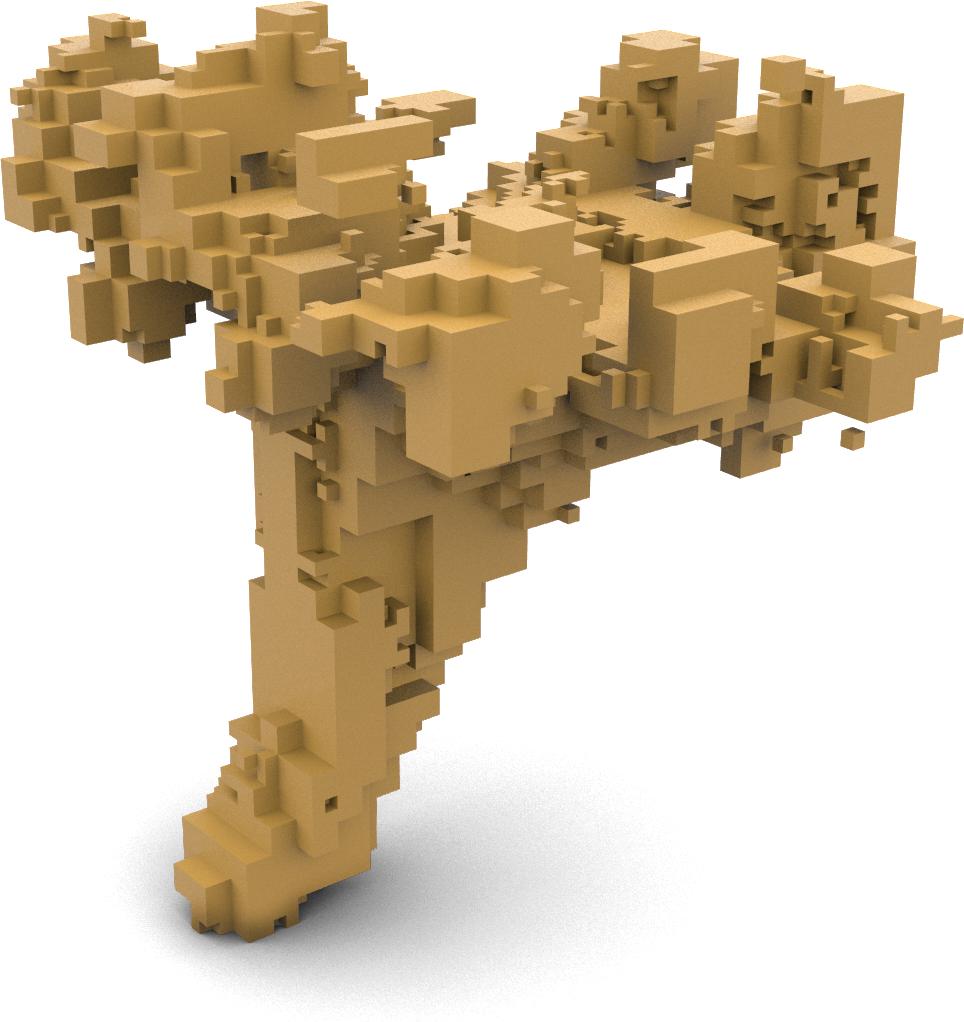}
    \caption{Chanterelle mushroom dataset samples and generated shapes from our model trained on this dataset.}
    \label{fig:mushrooms}
\end{figure}


\paragraph{Training on Natural Images}
Finally, we demonstrate that the proposed method is able to produce realistic samples when trained on a dataset of natural images. Figure \ref{fig:mushrooms} shows samples from a model trained on the Chanterelle mushrooms dataset from~\citet{platosCave}.

\subsection{Ablations}

\begin{figure}
    \centering
    \captionsetup[subfigure]{justification=centering}
    \begin{subfigure}[t]{.09\textwidth}
    \includegraphics[width=\textwidth]{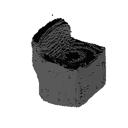}
    \includegraphics[width=\textwidth]{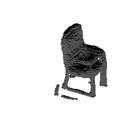}
    \caption{\\ 0.1}
    \end{subfigure}
    \begin{subfigure}[t]{.09\textwidth}
    \includegraphics[width=\textwidth]{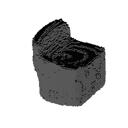}
    \includegraphics[width=\textwidth]{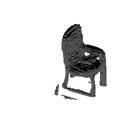}
    \caption{\\ 0.2}
    \end{subfigure}
    \begin{subfigure}[t]{.09\textwidth}
    \includegraphics[width=\textwidth]{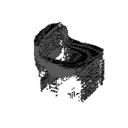}
    \includegraphics[width=\textwidth]{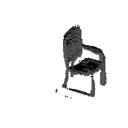}
    \caption{\\ 0.5}
    \end{subfigure}
    \begin{subfigure}[t]{.09\textwidth}
    \includegraphics[width=\textwidth]{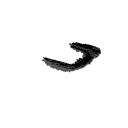}
    \includegraphics[width=\textwidth]{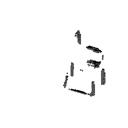}
    \caption{\\ 0.8}
    \end{subfigure}
    \begin{subfigure}[t]{.09\textwidth}
    \includegraphics[width=\textwidth]{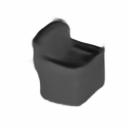}
    \includegraphics[width=\textwidth]{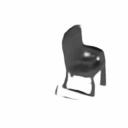}
    \caption{neural renderer}
    \end{subfigure}
\caption{Samples from a generator trained without DOM, rendered using the neural renderer on the continuous sample and OpenGL on various thresholds.}
\label{fig:Ablation}
\end{figure}

\paragraph{Discriminator output matching} We study the effect of the proposed discriminator output matching (DOM) loss in various scenarios. In Table \ref{tbl:Ablation}, we report the FID scores on the models trained without the DOM loss, from this comparison we see that the DOM loss plays a crucial role in learning to generate high-quality 3D objects. In Figure \ref{fig:Ablation} we can see that that the non-binary volumes sampled from the generator 
can be rendered to a variety of different images by OpenGL, depending on the random choice of threshold. Without the DOM loss, the trained neural renderer simply averages over these potential outcomes, considerably smoothing the result in the process and losing information about fine structures in the volume. This leads to weak gradients being passed to the generator, considerably deteriorating sample quality.
\begin{figure}[h]
    \captionof{table}{Ablation results without discriminator output matching (DOM) when training on \textcolor{red}{chairs}/\textcolor{blue}{couches} ``one per model'' datasets. We either fix the pre-trained neural renderer (``Fixed''), or continuing to train it during GAN training (``Retrained''). The generator samples fed to the discriminator are rendered using either OpenGL or the neural renderer. For reference, our model is equivalent to the Retrained OpenGL setup with the addition of the DOM loss and achieves FID scores \textcolor{red}{20.7}/\textcolor{blue}{35.8}.}
    \centering
    \begin{tabular}{c  c  c}
    & OpenGL & RenderNet \\ 
     \midrule 
    Retrained & \textcolor{red}{86.4}/\textcolor{blue}{180.1} & \textcolor{red}{74.7}/\textcolor{blue}{144.8} \\ 
    Fixed & \textcolor{red}{113.7}/\textcolor{blue}{323.5} & \textcolor{red}{103.9}/\textcolor{blue}{124.6} \\ 
    \end{tabular}
    \label{tbl:Ablation}
\end{figure}

Another setting from Table~\ref{tbl:Ablation} shows the discriminator trained using generated samples rendered by the neural renderer instead of OpenGL. This inherently prevents the mode collapse observed in the above setting. However, it leads to the generator being forced by the discriminator to produce binary voxel maps early on in training. This seems to lead to the generator getting stuck in a local optima, hence deteriorating sample quality.

\paragraph{Pre-training} We investigate the effect of various pre-trainings of the neural renderer. All other experiments were conducted with the neural renderer pre-trained on the \textit{Tables} data from ShapeNet (see Table \ref{tbl:maintable}). As a comparison, we run the proposed algorithm on the chair data using a neural renderer pre-trained on either the Chair data itself or a simple data set consisting of randomly sampled cubes. As shown in Table \ref{tbl:Pretraining}, the quality of the results produced by our method is robust to changes in the pre-training of the neural renderer. In contrast, if we use a fixed pre-trained renderer it produces reasonable results if pre-trained directly on the domain of interest, but deteriorates significantly if trained only on a related domain.  Note  that we assume no access to 3D data in the domain of interest so in practice we cannot pre-train in this way.

\begin{figure}[t]
    \captionof{table}{Comparisons of neural renderer pre-trainings on different 3D shapes. FIDs are reported for the 'One per model' chairs.}
    \centering
    \begin{tabular}{cccc}
     & Chairs & Tables & Random \\ 
     \midrule
    Ours & 22.6 & 20.7 & 20.4 \\ 
    Fixed & 37.8 & 105.8 & 141.9  \\ 
    \end{tabular}
    \label{tbl:Pretraining}
\end{figure}

~\section{Conclusion}
We have presented the first scalable algorithm using an arbitrary off-the-shelf renderer to learn to generate 3D shapes from unstructured 2D data. We have introduced a novel loss term, Discriminator Output Matching, that allows stably training our model on a variety of datasets, thus achieving significantly better FID scores than previous work. In particular, Using light exposure and shadow information in our rendering engine, we are able to generate high-quality convex shapes, like bathtubs and couches, that prior work failed to capture.

The presented framework is general and in-concept can be make to work with any off-the-shelf renderer. In practice, our work can be extend by using more sophisticated photo-realistic rendering engines, to be able to learn even more detailed information about the 3D world from images. By incorporating color, material and lighting prediction into our model we hope to be able to extend it to work with more general real world datasets.

\bibliography{main}
\bibliographystyle{icml2020}

\end{document}